\documentclass[11pt]{article}

\usepackage[preprint]{acl}

\usepackage{times}
\usepackage{latexsym}

\usepackage[T1]{fontenc}

\usepackage[utf8]{inputenc}

\usepackage{microtype}

\usepackage{inconsolata}

\usepackage{graphicx}

\usepackage[table]{xcolor}    
\usepackage{colortbl}         
\usepackage{booktabs}         

\usepackage{pgf}             
\usepackage{multirow}
\usepackage{tabularx}
\usepackage{arydshln}
\usepackage{gradient}
\usepackage{twemojis}

\usepackage{tikz}
\usetikzlibrary{shapes,snakes}
\usepackage{amsmath,amssymb}
\usepackage[inline]{enumitem}

\usepackage{booktabs}

\usepackage{pgfplots}
\pgfplotsset{compat=1.18}
\usetikzlibrary{patterns}

\newcommand{\Emoji}[2][1em]{\raisebox{-0.15ex}{\twemoji[height=#1]{#2}}}
\title{Bring the Apple \Emoji{apple}, Not the Sofa \Emoji{couch and lamp}: Impact of Irrelevant Context \\ in Embodied AI Commands on VLA Models}

\author{
 \textbf{Daria Pugacheva}\thanks{
   \textbf{Correspondence:} \href{mailto:email@domain}{sedyakina.d@gmail.com}
 }, \quad
 \textbf{Andrey Moskalenko}, \quad
 \textbf{Denis Shepelev}, \quad
\\
 \textbf{Andrey Kuznetsov}, \quad
 \textbf{Vlad Shakhuro}, \quad
 \textbf{Elena Tutubalina}
 \\
}

\begin{document}
\maketitle
\begin{abstract}


Vision Language Action (VLA) models are widely used in Embodied AI, enabling robots to interpret and execute language instructions. However, their robustness to natural language variability in real-world scenarios has not been thoroughly investigated.
In this work, we present a novel systematic study of the robustness of state-of-the-art VLA models under linguistic perturbations. 
Specifically, we evaluate model performance under two types of instruction noise: (1) human-generated paraphrasing and (2) the addition of irrelevant context.
We further categorize irrelevant contexts into two groups according to their length and their semantic and lexical proximity to robot commands.
In this study, we observe consistent performance degradation as context size expands. We also demonstrate that the model can exhibit relative robustness to random context, with a performance drop within 10\%, while semantically and lexically similar context of the same length can trigger a quality decline of around 50\%. Human paraphrases of instructions lead to a drop of nearly 20\%. 
To mitigate this, we propose an LLM-based filtering framework that extracts core commands from noisy inputs.
Incorporating our filtering step allows models to recover up to 98.5\% of their original performance under noisy conditions.

\end{abstract}


\section{Introduction}
\label{sec:introduction}

Embodied AI is undergoing rapid development, with robotic systems increasingly exhibiting practical utility in everyday environments.
Vision-Language-Action (VLA) models play a central role in enabling this progress.
By leveraging large language models (LLMs), robots can interpret and execute natural language instructions grounded in visual perception~\cite{open_x_embodiment_rt_x_2023, jiang2023vima, driess2023palm, zhou2025opendrivevla}.


\begin{figure}[t]
\centering
  \includegraphics[width=1.0\linewidth]{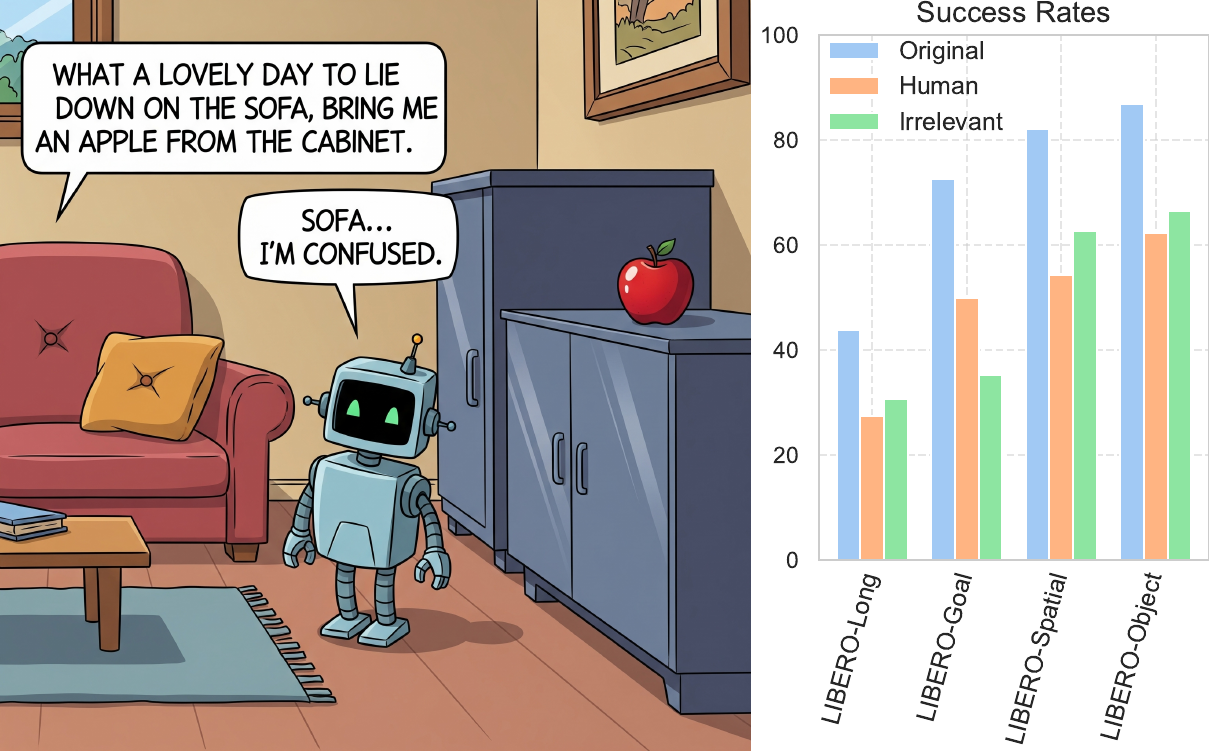}
  \caption{Human-voiced commands to the robot may contain irrelevant context and cause the target command to fail. We observed a significant drop in the success rates of VLA robotic models when real users posed problems.}
  \label{fig:confused-robot}
\end{figure}

Even when commands include irrelevant context or are paraphrased, which can occur in real-world human-robot communication (Figure~\ref{fig:confused-robot}), the performance of VLA models is expected to remain consistent.
However, the influence of linguistic variability on model performance remains insufficiently explored. 
For example, \citet{szot2024large} investigate the robustness of their proposed model to paraphrasing and irrelevant context, but their analysis is restricted to a limited set of templates, i.e. one for irrelevant context and four for paraphrasing.
Similarly, \citet{parekh-etal-2024-investigating} focus solely on template-based paraphrasing, but does not examine the influence of irrelevant context. 
Moreover, this work does not consider how real users might naturally paraphrase task instructions.

To address these gaps, we propose a novel set of instruction perturbations.
First, we develop an extensive range of irrelevant context types, including (1) contexts varying in length to assess the impact of irrelevant context length, and (2) contexts  based on their semantic and lexical proximity to commands from the training set of the VLA model.
Second, we collect human-generated paraphrases for all considered robot instructions to study the effects of natural language variation.

We perform evaluations using two well-known simulation benchmarks, LIBERO~\cite{liu2023libero} and Habitat 2.0~\cite{szot2021habitat}.
Our study covers five state-of-the-art VLA models: OpenVLA~\cite{pmlr-v270-kim25c}, UniAct~\cite{zheng2025universal}, MoDE~\cite{reuss2025efficient}, 
$\pi_0$~\cite{black2024pi_0},
and LLARP~\cite{szot2024large}.

Overall, our contributions are as follows:
\begin{itemize}
    \item We evaluate existing modern VLA models for various embodiments and identify that these models are most vulnerable to irrelevant context, which is lexically and semantically close to the commands from the training set.
    Moreover, we show that the performance degrades as the length of irrelevant context increases and can drop by up to 58\%, when the context length approaches the length of target commands.
    \item We perform a human study and show that natural paraphrasing drops VLA model performance by 20\%, revealing adaptation gaps between LLM-based VLA models and real-world deployment needs.
    \item We propose a filtering framework to preprocess noisy commands, which employs LLM to remove irrelevant context. This framework significantly enhances robot's execution robustness and improves success rates.
\end{itemize}



\section{Related Works}
\label{sec:related}

Vision-Language-Action (VLA) models enable robots to take visual observations and natural language commands as input and output low-level actions for control. We focused on the task of assessing robustness of these models to linguistic variation – the ability to understand paraphrased or syntactically altered commands that were not seen during training, which is crucial for real-world applications of VLA models.

\subsection{VLA Models}

Recent advances in VLA models have demonstrated the integration of web-scale multimodal pretraining with robotic control through co-fine-tuning of vision–language models on robot trajectory datasets. 

RT-1~\cite{rt12022arxiv} was a pioneering VLA-like model for real-world robotic manipulation. It processes a short sequence of camera images together with a task description in natural language, and outputs a sequence of robot actions. 
RT-2~\cite{rt22023arxiv} exhibits emergent semantic reasoning and generalization to novel objects and instructions by encoding actions as text tokens alongside natural language. 

Significant progress in the field has occurred with the release of the open-source OpenVLA~\cite{pmlr-v270-kim25c} foundation model, which explicitly integrates a large language model to strengthen language understanding. 
OpenVLA is a 7B policy built on a Llama 2~\cite{touvron2023llama} model, fused with vision encoders for image input. 
It was trained on 970k real robot demonstrations~\cite{open_x_embodiment_rt_x_2023} drawn from diverse sources, as well as additional “Internet-scale” vision-language data to inject world knowledge. Due to its openness, this model became the basis for subsequent work in this area~\cite{black2024pi_0, belkhale2024minivla, wen2025tinyvla, qu2025spatialvla, zheng2025universal, reuss2025efficient, lykov2025cognitivedrone}. 
Moreover, this approach was also utilized in drone control~\cite{lykov2025cognitivedrone, serpiva2025racevla} and autonomous vehicles~\cite{arai2025covla, zhou2025opendrivevla}. 
Thus, due to the significant growth of popularity of the models of the VLA family, we are conducting our research to understand the robustness of such models to the variability of text prompts. 


\subsection{Evaluation in Simulation Environments}

As a rule, robotics models are usually evaluated using success rate (SR) in a simulator and real world environments.
We believe it would be unsafe to evaluate deviant robotic behavior in the real world, so we focus mainly on simulator environments.
Unlike the real world, simulators allow accurate reproduction of all initial states, so different models can be compared objectively.
Thus, simulation environments have become indispensable for systematically benchmarking robotics models under controlled yet diverse conditions.

There are many simulation environments available. RoboCasa~\cite{robocasa2024} is a simulation framework for training generalist robots in realistic home environments. SimplerENV~\cite{li24simpler} offers a suite of simulated replicas of common real-robot setups, enabling scalable, reproducible evaluation and demonstrating strong correlation with real-world performance for generalist policies. 

Habitat~\cite{savva2019habitat} is a high-performance simulator for embodied AI and navigation tasks, capable of rendering RGB-D observations and simulating rigid-body dynamics at over 8,000 steps per second in photorealistic 3D scenes.



LIBERO~\cite{liu2023libero} provides a lifelong learning benchmark with procedurally generated manipulation tasks, specifically designed to study declarative and procedural knowledge transfer in simulation at scale. LIBERO is organized into four distinct task suites designed to probe different facets of lifelong learning in robot manipulation.

We mainly focused on Habitat and LIBERO for our experiments, since they are now popular simulation environments to benchmark VLA models.

\subsection{VLA Robustness}

Robustness is an active area of evaluation for VLA models. Recent comparative studies have explicitly tested a number of models on paraphrased or altered instructions to probe their robustness. LADEV~\cite{wang2024ladev} is a language-driven evaluation framework that generates paraphrases of task instructions (using LLMs generation method) to test VLA policies. Researchers compared multiple models on the same set of tasks under original and paraphrased commands. We extend this research by using a simulator with a larger number of robotic tasks, as well as we also proposed intelligent generation of text paraphrases of different categories, and also showed how to improve the robustness of models to such reformulations.

\citet{wang2024exploring} presented a study of adversarial attacks on Vision-Language-Action models, highlighting novel vulnerabilities unique to robotic control tasks. They introduce two attack objectives: an untargeted position-aware attack that perturbs spatial inputs to destabilize controller outputs and a targeted manipulation attack that 
crafts minimal perturbations to redirect robot trajectories toward specific failure modes.  However, the authors study only image perturbation robustness at the robot's input, which is a rarer case because the robot's camera is inside it and can only be attacked with physically printed patches. We study resistance specifically to text prompts because the user always has direct influence on them.

\section{Evaluating VLA Models}
\label{sec:eval_vla_models}

\begin{table*}[!ht]
\centering
\scalebox{0.97}{
\begin{tabular}{c|ll|p{0.6\textwidth}}
\hline
\textbf{Environment} & \multicolumn{2}{c|}{\textbf{Variation}} & \textbf{Command} \\
\hline

 \multirow{8}{*}{\begin{tabular}[c]{@{}c@{}}\\Habitat 2.0\\ \includegraphics[width=0.15\linewidth]{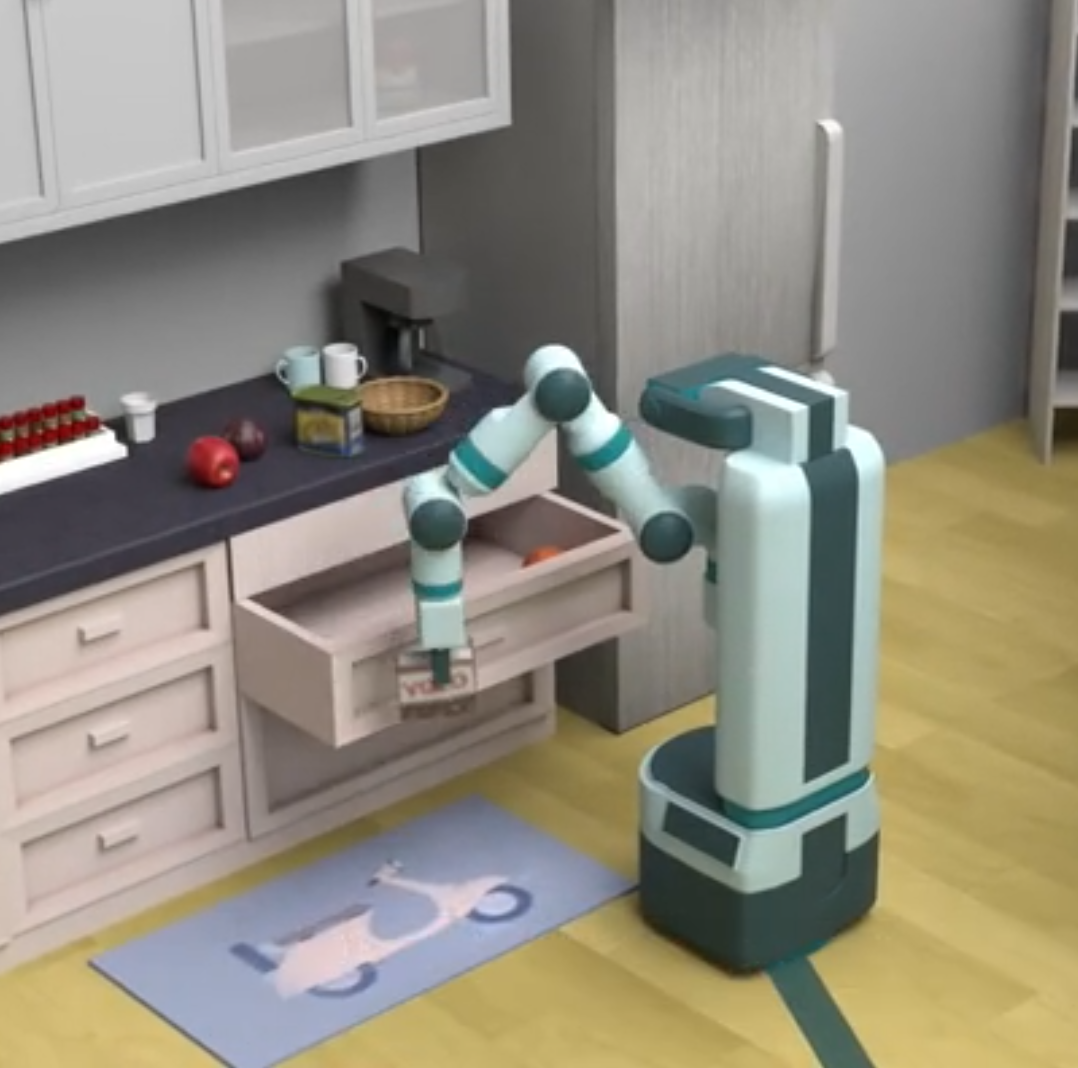}\end{tabular}} &  & Original & Find an orange and move it to the sink. \\
  \cline{2-4}
    &  & Human & Can you find an orange and put it in the sink? \\
\cline{2-4}
   & \multirow{4}{*}{\centering\rotatebox[origin=c]{90}{\textbf{Context}} \vspace{0.1pt} \centering\rotatebox[origin=c]{90}{\textbf{Length}}} & Single & \textit{Although,} find an orange and move it to the sink. \\
 & & Short & \textit{Inspired while cooking dinner.} Find an orange and move it to the tv stand. \\
    & & Long & \textit{He felt motivated cleaning the pantry and organizing everything, so} find an orange and move it to the sink. \\
\cline{2-4}
    & \multirow{4}{*}{\centering\rotatebox[origin=c]{90}{\textbf{Context}} \vspace{0.1pt} \centering\rotatebox[origin=c]{90}{\textbf{Semantic}}} & Location & \textit{There's an apple on the TV stand, but} find an orange and move it to the sink. \\
    & & Description & \textit{Cup is a container for liquids.} Find an orange and move it to the sink. \\
    & & Infeasible & \textit{Bake a pie with peach slices.} Find an orange and move it to the sink. \\
\hline
\multirow{8}{*}{\begin{tabular}[c]{@{}c@{}}\\LIBERO\\ \includegraphics[width=0.15\linewidth]{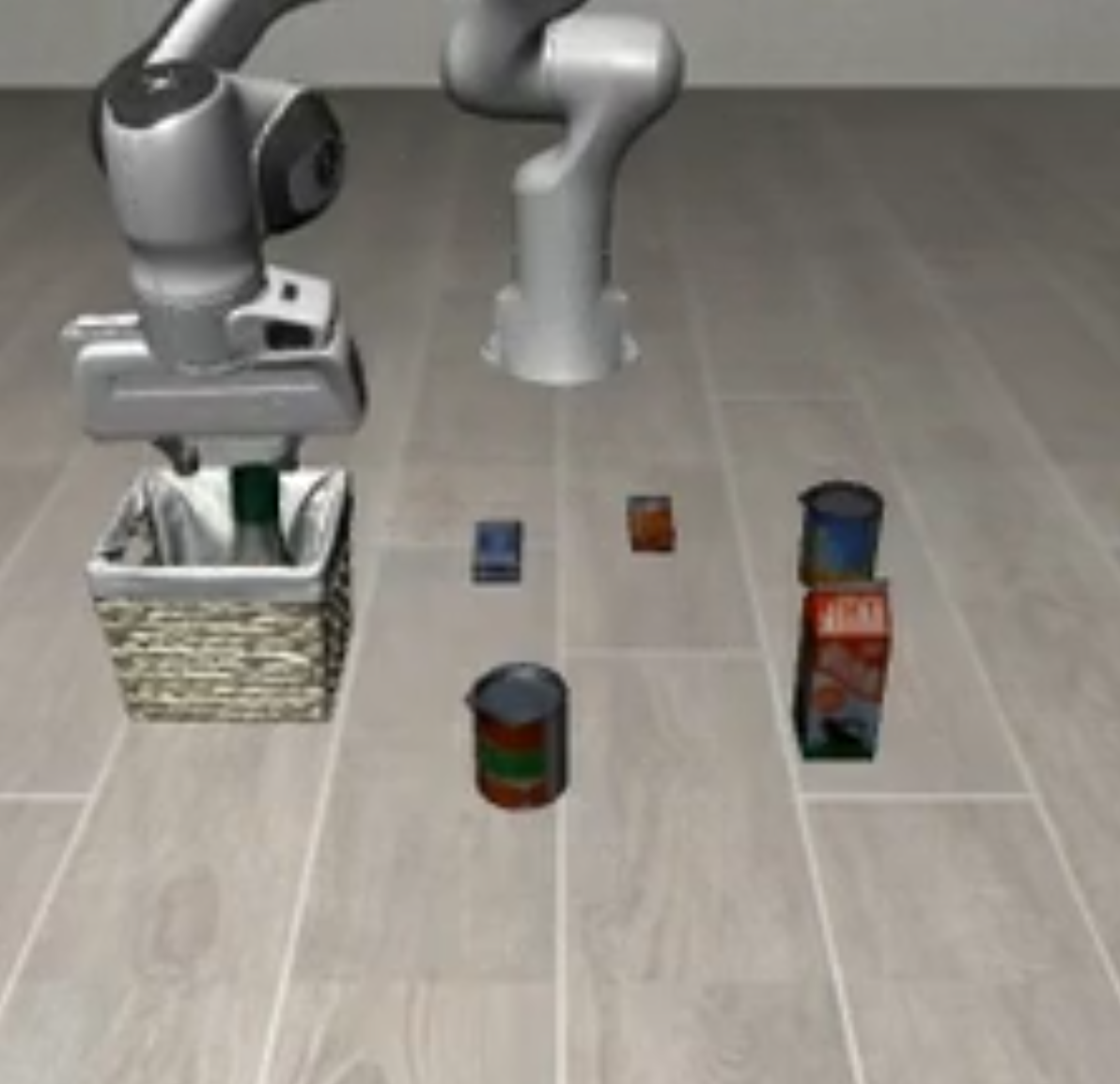}\end{tabular}}
    & & Original & put the wine bottle on top of the cabinet \\
    \cline{2-4}
    &  & Human & move the bottle of wine to the top of the cabinet \\
\cline{2-4}
    & \multirow{4}{*}{\centering\rotatebox[origin=c]{90}{\textbf{Context}} \vspace{0.1pt} \centering\rotatebox[origin=c]{90}{\textbf{Length}}} & Single & \textit{moreover} put the wine bottle on top of the cabinet \\
 & & Short & \textit{nostalgia strikes after dinner} put the wine bottle on top of the cabinet \\
 & & Long & \textit{the gloomy weather matched her tired and melancholy} put the wine bottle on top of the cabinet \\
\cline{2-4}
    & \multirow{4}{*}{\centering\rotatebox[origin=c]{90}{\textbf{Context}} \vspace{0.1pt} \centering\rotatebox[origin=c]{90}{\textbf{Semantic}}} & Location & \textit{the bowl is in the basket} put the wine bottle on top of the cabinet \\
    & & Description & \textit{padlock are made of metal} put the wine bottle on top of the cabinet \\
    & & Infeasible & \textit{bite into the soft plum} put the wine bottle on top of the cabinet \\    
\hline
\end{tabular}
}
\caption{Examples of context inserted into commands for the Habitat 2.0 simulator and LIBERO benchmark.}
\label{tab:context_example}
\end{table*}


In this section, we describe the evaluation setup for the VLA models.
We begin with introducing the simulation environments~\ref{sec:env} and VLA models~\ref{sec:vla_models} used in our study.
Next, we propose several types of irrelevant context~\ref{sec:irrelevant_content} and present crowdsourced paraphrases of robot commands~\ref{sec:experiments_human} to assess model robustness.
Finally, we report experimental results and provide their analysis~\ref{sec:experiments}.



\subsection{Simulation Environments}
\label{sec:env}

We study the robustness of the VLA models in the LIBERO~\cite{liu2023libero} and Habitat 2.0~\cite{szot2021habitat} simulation environments.

LIBERO~\cite{liu2023libero} is designed to evaluate models on object manipulation tasks.
Each LIBERO task suite focuses on a specific type of distribution shift or knowledge transfer challenge, enabling controlled evaluation of model capabilities under spatial, object, goal, and entangled task variations.
We consider the following LIBERO task suites:
\begin{itemize}
\item  \textit{LIBERO-Spatial:} contains 10 short-horizon tasks that require the robot to transfer and memorize new spatial relationships.

\item \textit{LIBERO-Object}: comprises 10 short-horizon tasks centered on learning new object types, where the robot must pick and place different objects in sequence.

\item \textit{LIBERO-Goal:} includes 10 short-horizon tasks that share identical objects and spatial layouts but differ only in procedural goals, testing the transfer of motion and behavior knowledge.

\item \textit{LIBERO-Long} (also called LIBERO-10) comprises 10 long-horizon tasks, reserved for downstream evaluation of lifelong learning algorithms.
\end{itemize}

Habitat 2.0~\cite{szot2021habitat} is a simulation platform that supports not only object manipulation but also navigation tasks. Following the authors' instructions~\cite{szot2024large}, we generated 100 language commands for evaluation. 
Both the generated commands and those from the training set included punctuation marks and letters in various cases, such as, ``Find an apple and put it away in the fridge.'' Moreover, these commands could also be phrased as questions, offering a greater diversity compared to the commands found in LIBERO.

\subsection{VLA Models}
\label{sec:vla_models}

In LIBERO, we evaluate three state-of-the-art and popular models: OpenVLA~\cite{pmlr-v270-kim25c}, UniAct~\cite{zheng2025universal}, Mixture-of-Denoising Experts (MoDE)~\cite{reuss2025efficient}, $\pi_0$~\cite{black2024pi_0}. 
In Habitat 2.0, we evaluate LLARP model~\cite{szot2024large}.

To ensure lower variance in the experimental results, models are evaluated on LIBERO benchmarks across 50 trials for each task suite, and the reported performance is the average success rate over three random seeds (resulting in 150 total trials per statistic).

During the rollout phase of LLARP, the policy acts in parallel in 32 Habitat 2.0 environments and are evaluated across 30 trials for each task, and the reported performance is the average success rate over three random seeds as well.

\subsection{Irrelevant Context}
\label{sec:irrelevant_content}

We consider several types of irrelevant context and organize them into two groups: (1) context length variation, (2) semantic and lexical similarity.

The first group of contexts was chosen to be lexically and semantically different from the commands of the training set, and varied in length. The context from the second group contained names of scene objects and constructions similar to the training commands. 
All contexts are generated using GPT 4.1 and then verified by experts.
Each context is added both before the target command and afterward.
We adapt the final noisy command to maximize similarity to the template from the model training set in order to eliminate the possible impact of punctuation and letter case changes (please see Tab.~\ref{tab:context_example} with examples). 

\paragraph{Context length variation}
Specifically, the first set consists of a context \textit{``Single''}, which includes single introductory word like `However', 'Moreover' etc; contexts \textit{``Short''} and \textit{`Long'} includes 3-5  or 7-10 words sentences whose content represented random phrases unrelated to the roboarm commands or objects in the scene, e.g., `the weather is nice today' or `the gloomy weather matched her tired and melancholy mood today'. 

\paragraph{Semantic and lexical similarity}
The second set also comprises three types of context. 

The first type of context \textit{``Description''} provides semantic proximity to the training set. It contains short phrases describing the random object of the scene, but this description was arbitrary. It did not include information about the location of the object or the action to be performed with the object, e.g. ``Cup is a container for liquids. Find an orange and move it to the TV stand''. 

The next type \textit{``Infeasible''} represents infeasible commands, which the roboarm cannot execute, and which did not occur in the training set, e.g., ``Bake a pie with peach slices. Find an orange and move it to the right counter''.
It is semantically and grammatically close to training commands, but differs lexically.

Finally, the last type \textit{``Location''} combines both semantic and lexical proximity to what the model observed in training. It consists of short phrases with 3-5 words that contain references to the location of the objects in the scene. The location and the names of the objects themselves did correspond to the content of the scene, but the subsequent command was not related to the object, e.g., `There's an apple in the cabinet, but find a screwdriver and move it to the left counter.'  
A more complete list of examples for each type of context can be found in Appendix.

For each target command, context was injected both before and after the command. We provide averaged results for these two injection types.

\subsection{Command Paraphrasing}
\label{sec:experiments_human}

To evaluate the robustness of VLA models to command paraphrasing, we conducted a real-user study.
Specifically, crowdworkers were asked to paraphrase task descriptions drawn from experimental simulation benchmarks.
All commands were originally written in English, so we restricted participation to workers who passed an English-proficiency test.
To avoid introducing annotation bias, instructions were kept as minimal as possible, with the sole requirement that the reformulated text preserve the meaning of the original.
Participants saw the instruction from Figure~\ref{fig:crowd}.

Each worker received a batch of five descriptions per task and spent on median 296 seconds (including instruction time) to complete the task. Each description was independently paraphrased by five different crowdworkers.

All collected paraphrases were then reviewed by our in-lab experts, who retained only those submissions in which the semantic content of the original description was faithfully preserved.

The resulting texts were then used to evaluate the performance of the VLA models by replacing the original task prompts in the simulation benchmarks with texts formulated by real-users.


  \tikzstyle{mybox} = [draw=gray!50, fill=gray!7, very thick,
    rectangle, rounded corners, inner sep=10pt, inner ysep=10pt]
  \tikzstyle{fancytitle} = [draw=gray!50, very thick, fill=white]
  \tikzstyle{fancyorig} = [draw=gray!50, fill=gray!7, very thick]

\begin{figure}[!hbt]
\centering
\begin{tikzpicture}[baseline=-2cm]
    \node [mybox] (box){
      \begin{minipage}[t!]{0.4\textwidth}
      
Thank you for taking part  \\
in the experiment!
\hfill \break

Imagine that you want to assign a task to a physical \textbf{robot}. The original task is written in text in the "\textbf{Original Text}" field.
\hfill \break

You are required to \textbf{rephrase} this text and write it in your own words in the ``\textbf{Your rephrasing}'' (in your own words) field, keeping the original meaning.
\end{minipage}
};
\node[fancytitle, text width=0.43\textwidth, text centered, rounded corners] at (box.north) {Crowdsourcing participants instruction};
\end{tikzpicture}
\caption{The instruction that was shown to workers during crowdsourcing.}
\label{fig:crowd}
\end{figure}
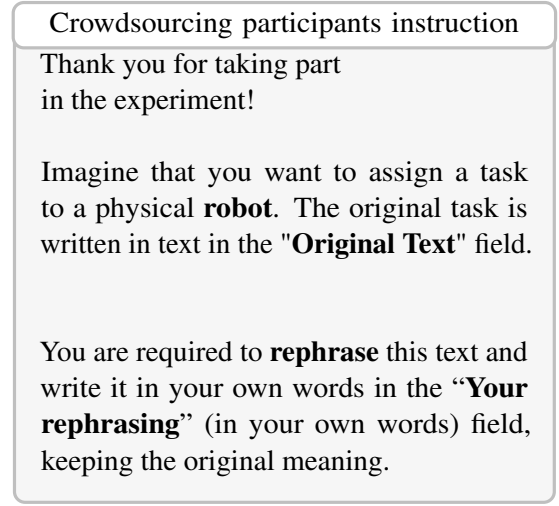

\subsection{Results and Analysis}
\label{sec:experiments}

\begin{table*}[!htbp]
\centering
\fontsize{11pt}{14pt}\selectfont
\tabcolsep=2.55pt
\scalebox{0.88}{
\begin{tabular}{ccc|ccc|ccc|cc}
\toprule
\multirow{2}{*}{\textbf{Environment}} & \multirow{2}{*}{\textbf{Model}} & \multirow{2}{*}{\textbf{Original}} & \multicolumn{3}{c|}{\textbf{Length}} & \multicolumn{3}{c|}{\textbf{Semantic}} & \multicolumn{2}{c}{\textbf{Paraphrasing}} \\ \cline{4-11}
& & & \textbf{Single} & \textbf{Short} & \textbf{Long} & \textbf{Description} & \textbf{Infeasible} & \textbf{Location} & \textbf{Human} & \textbf{DeepSeek} \\ \hline
\multirow{3}{*}{\begin{tabular}[c]{@{}c@{}}LIBERO\\ Goal\end{tabular}} &
OpenVLA &
\gradientnums{18.9}{77.5}{77.5} &
\gradientnums{18.9}{67.6}{67.6} &
\gradientnums{18.9}{67.6}{43.5} &
\textbf{\gradientnums{18.9}{67.6}{18.9}} &
\gradientnums{25.5}{30.4}{30.4} &
\gradientnums{25.5}{30.4}{28.0} &
\textbf{\gradientnums{25.5}{30.4}{25.5}} &
\gradientnums{54.8}{58.2}{58.2} &
\textbf{\gradientnums{54.9}{54.8}{54.8}} \\
& UniAct &
\gradientnums{16.0}{67.5}{67.5} &
\gradientnums{28.5}{62.5}{62.5} &
\gradientnums{28.5}{62.5}{39.5} &
\textbf{\gradientnums{28.5}{62.5}{28.5}} &
\gradientnums{16.0}{30.5}{30.5} &
\gradientnums{16.0}{30.5}{28.3} &
\textbf{\gradientnums{16.0}{30.5}{16.0}} &
\gradientnums{38.8}{41.7}{41.7} &
\textbf{\gradientnums{38.8}{41.7}{38.8}} \\ 
& $\pi_0$ &
\gradientnums{47.0}{91.5}{91.5} &
\gradientnums{44.8}{91.6}{91.6} &
\gradientnums{44.8}{91.6}{77.9} &
\textbf{\gradientnums{44.8}{91.6}{44.8}} &
\gradientnums{55.6}{68.8}{68.8} &
\gradientnums{55.6}{68.8}{59.6} &
\textbf{\gradientnums{55.6}{68.8}{55.6}} &
\gradientnums{71.2}{78.5}{78.5} &
\textbf{\gradientnums{71.2}{78.5}{71.2}} \\ 
\hline
\multirow{3}{*}{\begin{tabular}[c]{@{}c@{}}LIBERO\\ Object\end{tabular}} &
OpenVLA &
\gradientnums{56.2}{87.3}{87.3} &
\gradientnums{56.2}{86.3}{86.3} &
\gradientnums{56.2}{86.3}{74.2} &
\textbf{\gradientnums{56.2}{86.3}{56.2}} &
\gradientnums{62.5}{72.5}{70.3} &
\textbf{\gradientnums{62.5}{72.5}{62.5}} &
\gradientnums{62.5}{72.5}{72.5} &
\textbf{\gradientnums{80.0}{82.8}{80.0}} &
\gradientnums{80.0}{82.8}{82.8} \\
& UniAct &
\gradientnums{44.6}{86.5}{86.5} &
\gradientnums{47.0}{82.0}{82.0} &
\gradientnums{47.0}{82.0}{64.0} &
\textbf{\gradientnums{47.0}{82.0}{47.0}} &
\gradientnums{55.3}{63.8}{59.8} &
\textbf{\gradientnums{55.3}{63.8}{55.3}} &
\gradientnums{55.3}{63.8}{63.8} &
\textbf{\gradientnums{44.6}{58.2}{44.6}} &
\gradientnums{44.6}{58.2}{58.2}\\ 
& $\pi_0$ &
\gradientnums{83.5}{98}{97.5} &
\gradientnums{84.9}{97.4}{97.4} &
\gradientnums{84.9}{97.4}{94.8} &
\textbf{\gradientnums{84.9}{97.4}{84.9}} &
\gradientnums{85.9}{91.8}{91.8} &
\gradientnums{85.9}{91.8}{92.5} &
\textbf{\gradientnums{85.9}{91.8}{85.9}} &
\textbf{\gradientnums{89.7}{95.8}{89.7}} &
\gradientnums{89.7}{95.8}{95.8}\\ 
\hline
\multirow{3}{*}{\begin{tabular}[c]{@{}c@{}}LIBERO\\ Spatial\end{tabular}} &
OpenVLA &
\gradientnums{52.0}{85.3}{85.3} &
\gradientnums{52.0}{82.0}{82.0} &
\gradientnums{52.0}{82.0}{66.5} &
\textbf{\gradientnums{52.0}{82.0}{52.0}} &
\gradientnums{61.5}{62.5}{61.9} &
\textbf{\gradientnums{61.5}{62.5}{61.5}} &
\gradientnums{61.5}{62.5}{62.5} &
\textbf{\gradientnums{58.0}{64.1}{58.0}} &
\gradientnums{58.0}{64.1}{64.1} \\
& UniAct &
\gradientnums{50.5}{79.0}{79.0} &
\gradientnums{50.5}{69.8}{69.8} &
\gradientnums{50.5}{69.8}{61.5} &
\textbf{\gradientnums{50.5}{69.8}{50.5}} &
\textbf{\gradientnums{57.8}{61.5}{57.8}} &
\gradientnums{57.8}{61.5}{59.0} &
\gradientnums{57.8}{61.5}{61.5} &
\gradientnums{50.0}{50.5}{50.5} &
\textbf{\gradientnums{50.0}{50.5}{50.0}} \\ 
& $\pi_0$ &
\gradientnums{82.0}{97}{96.7} &
\gradientnums{76.9}{97.5}{97.5} &
\gradientnums{76.9}{97.5}{94.9} &
\textbf{\gradientnums{76.9}{97.5}{76.9}} &
\gradientnums{80.9}{92.5}{92.5} &
\gradientnums{80.9}{92.5}{88.5} &
\textbf{\gradientnums{80.9}{92.5}{80.9}} &
\textbf{\gradientnums{82.0}{97}{88.0}} &
\gradientnums{82.0}{97}{91.2}\\ 
\hline
\multirow{4}{*}{\begin{tabular}[c]{@{}c@{}}LIBERO\\ Long\end{tabular}} &
OpenVLA &
\gradientnums{25.0}{51.7}{51.7} &
\gradientnums{30.5}{48.5}{48.5} &
\gradientnums{30.5}{48.5}{32.8} &
\textbf{\gradientnums{30.5}{48.5}{30.5}} &
\gradientnums{23.5}{36.0}{36.0} &
\gradientnums{23.5}{36.0}{30.5} &
\textbf{\gradientnums{23.5}{36.0}{23.5}} &
\gradientnums{30.0}{36.0}{36.0} &
\textbf{\gradientnums{30.0}{36.0}{30.0}} \\
& UniAct &
\gradientnums{18.8}{46.5}{46.5} &
\gradientnums{21.8}{32.5}{32.5} &
\gradientnums{21.8}{32.5}{28.0} &
\textbf{\gradientnums{21.8}{32.5}{21.8}} &
\textbf{\gradientnums{25.3}{30.3}{25.3}} &
\textbf{\gradientnums{25.3}{30.3}{25.3}} &
\gradientnums{25.3}{30.3}{30.3} &
\gradientnums{15.2}{18.8}{18.8} &
\textbf{\gradientnums{15.2}{18.8}{15.2}} \\
& $\pi_0$ &
\gradientnums{66.5}{88}{88.5} &
\gradientnums{64.4}{84.6}{84.6} &
\gradientnums{64.4}{84.6}{78.9} &
\textbf{\gradientnums{64.4}{84.6}{64.4}} &
\gradientnums{73.0}{79.8}{78.9} &
\gradientnums{73.0}{79.8}{79.8} &
\textbf{\gradientnums{73.0}{79.8}{73.0}} &
\gradientnums{76.9}{79.3}{79.3} &
\textbf{\gradientnums{76.9}{79.3}{76.9}} \\ 
& MoDE &
\gradientnums{80.3}{95.5}{95.5} &
\gradientnums{80.3}{94.0}{94.0} &
\gradientnums{80.3}{94.0}{91.8} &
\textbf{\gradientnums{80.3}{94.0}{80.3}} &
\gradientnums{84.3}{87.5}{87.5} &
\gradientnums{84.3}{87.5}{85.0} &
\textbf{\gradientnums{84.3}{87.5}{84.3}} &
\gradientnums{80.3}{90.9}{90.9} &
\gradientnums{80.3}{95.5}{-}\\ \hline
Habitat 2.0 & LLARP &
\gradientnums{46.2}{98.3}{98.3} &
\gradientnums{60.7}{97.5}{97.5} &
\gradientnums{60.7}{97.5}{90.8} &
\textbf{\gradientnums{60.7}{97.5}{60.7}} &
\gradientnums{46.2}{89.8}{89.8} &
\gradientnums{46.2}{89.8}{57.8} &
\textbf{\gradientnums{46.2}{89.8}{46.2}} &
\textbf{\gradientnums{83.7}{97.4}{83.7}} &
\gradientnums{83.7}{97.4}{97.4}\\ \bottomrule
\end{tabular}
}
\caption{
    Success rates of VLA-models on different task suits and language perturbations. 
    Bold type indicates the largest drop in the success rate across each group of perturbations.
}
\label{tab:irrelevant}
\end{table*}

\begin{figure*}[t] 

    \centering 

    \begin{minipage}{0.19\textwidth}
        \includegraphics[width=\linewidth]{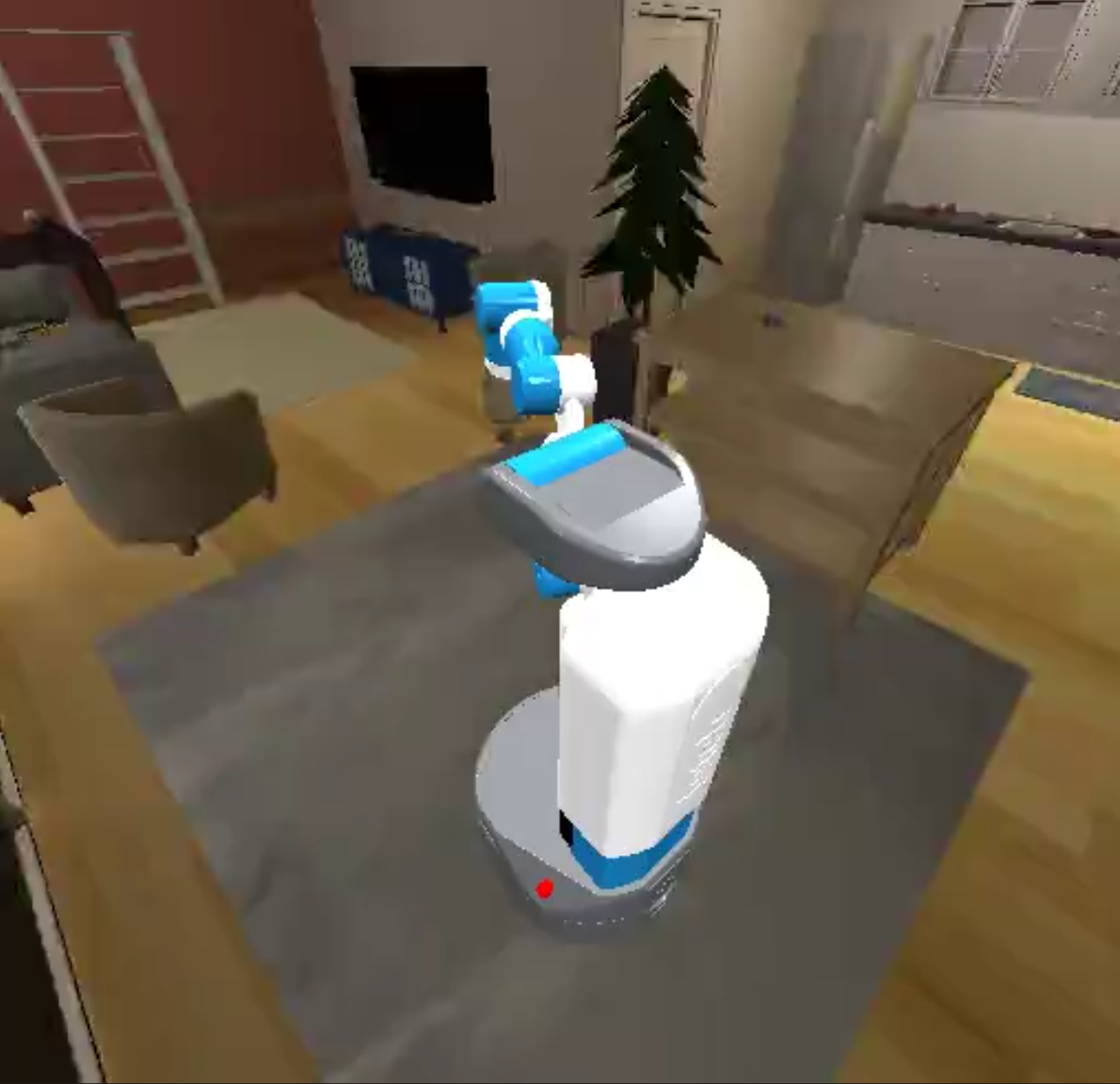}

        \caption*{Start scene: \\ navigate to sofa 
        \hfill \break
        
        } 

    \end{minipage}\hfill 
    \begin{minipage}{0.19\textwidth}
        \includegraphics[width=\linewidth]{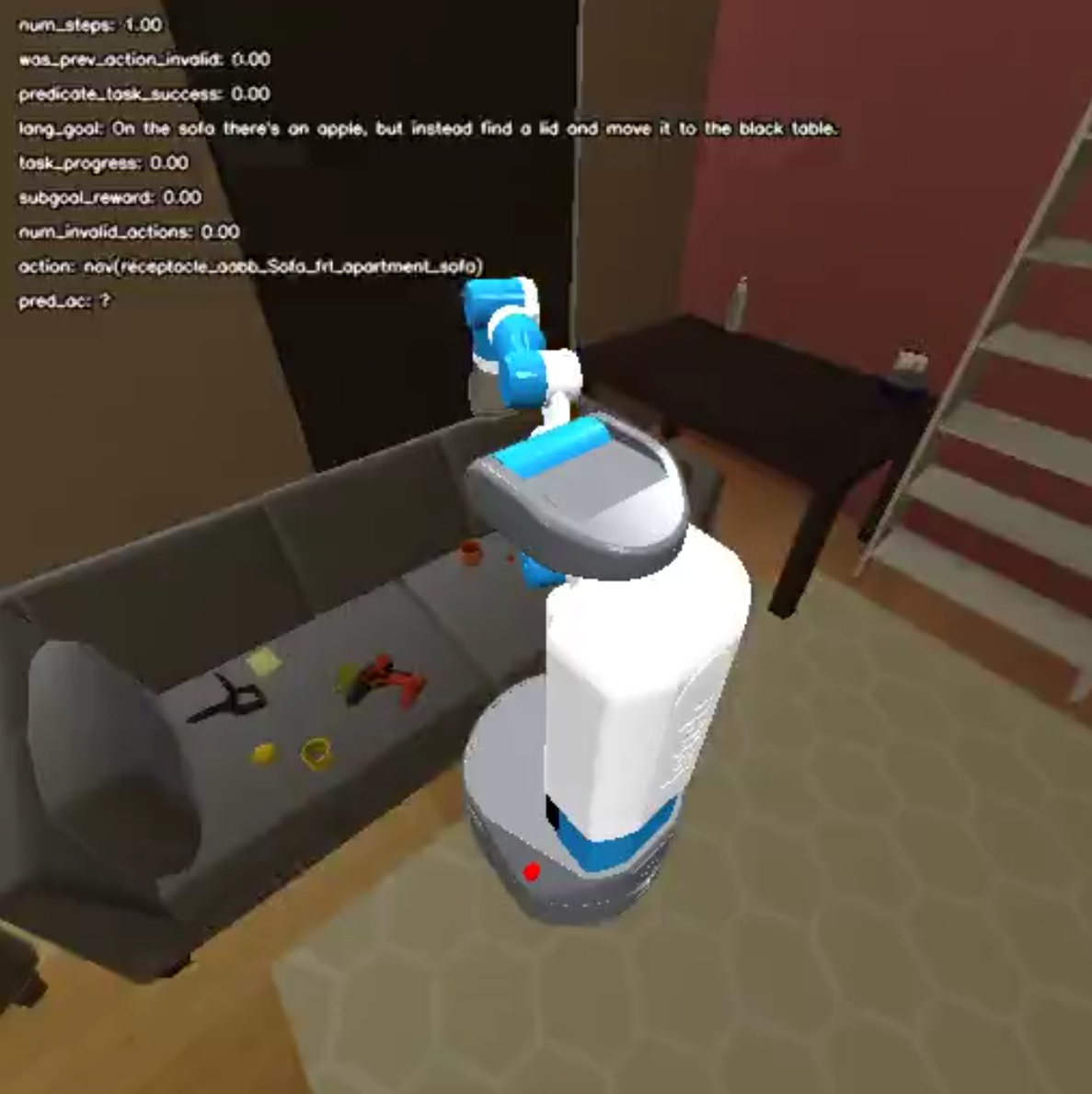}

        \caption*{Scene 2: pick lid, pick box, navigate to left counter}

    \end{minipage}\hfill
    \begin{minipage}{0.19\textwidth}

        \includegraphics[width=\linewidth]{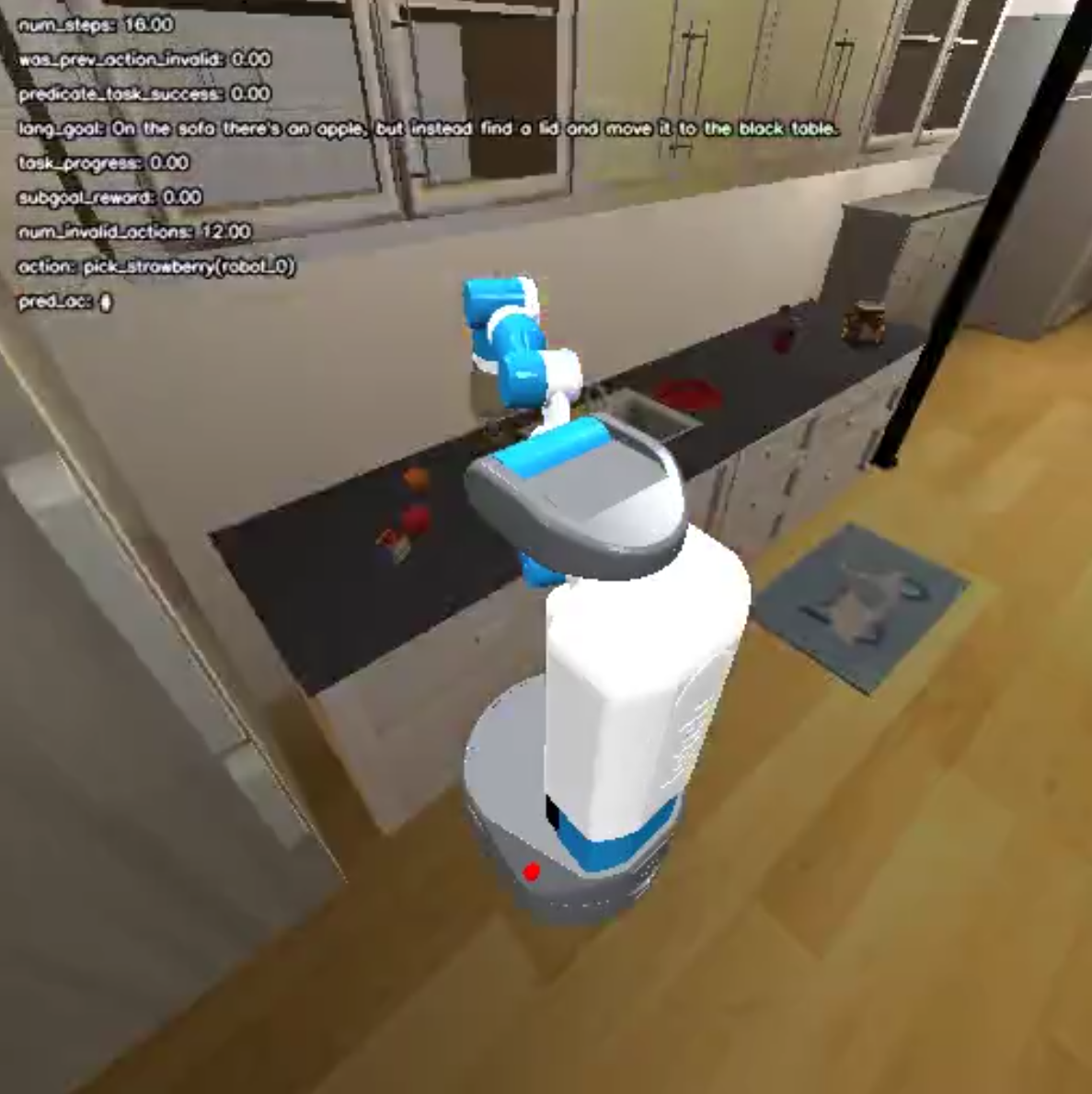}

        \caption*{Scene 3: pick lego, pick strawberry, navigate to brown table }

    \end{minipage}\hfill
    \begin{minipage}{0.19\textwidth}

\includegraphics[width=\linewidth]{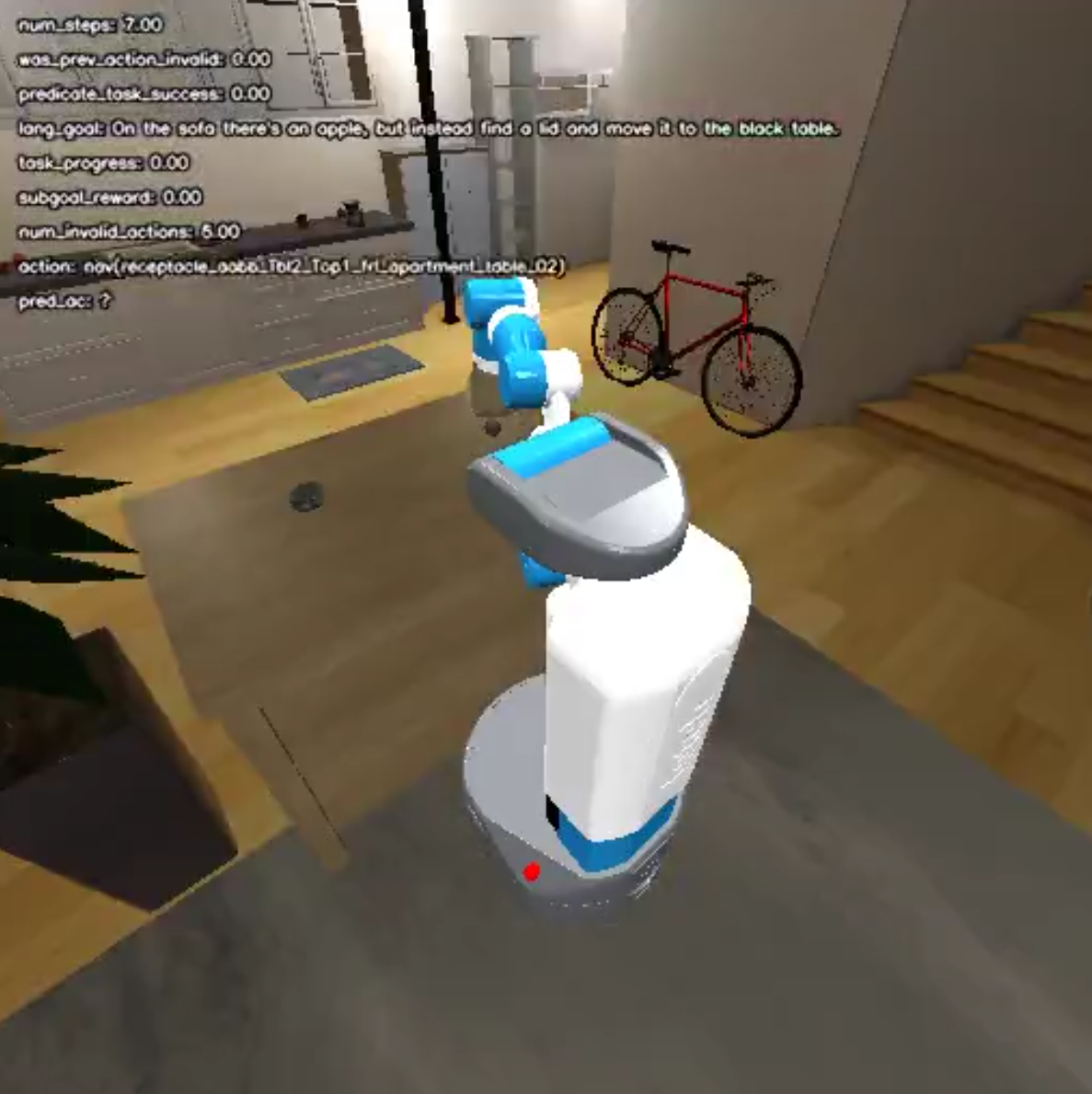} 

        \caption*{Scene 4: pick toy airplane, navigate to black table}

    \end{minipage}\hfill
    \begin{minipage}{0.19\textwidth}
    \includegraphics[width=\linewidth]{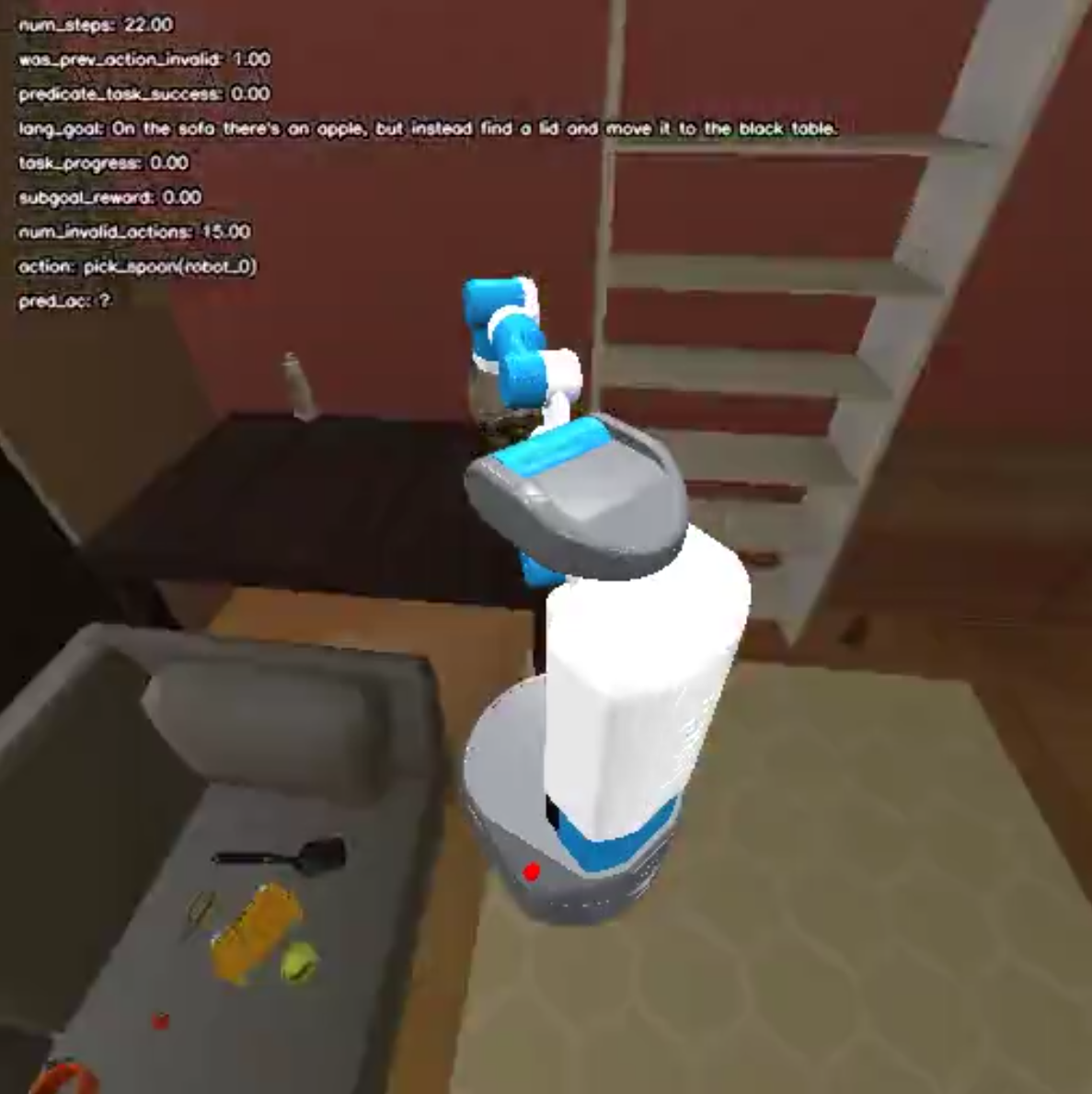} 

        \caption*{Final scene: \\ pick spoon 
        \hfill \break

        }

    \end{minipage}

    \caption{Demonstration of invalid robot behavior in a Habitat 2.0 simulator under the influence of irrelevant context ``On the sofa there's an apple'' for the target command ``find a lid and move it the black table''. The images correspond to the sequence of scenes from the episode. The captions under the scene images correspond to the actions that the robot executes.} 

    \label{fig:llarp-scenes} 

\end{figure*}

\paragraph{Command Paraphrasing by Human.}
According to the column ``Human'' in Table~\ref{tab:irrelevant}, natural command paraphrases lead to a lower number of successful episodes.
Workers tend to use different synonyms in language commands, the vocabulary used is larger, and people do not tend to stick to any pattern of language command construction. 
Natural noise entering language commands tended to be a few words long and often contained various words related to politeness such as, but not limited to, ``please'' and ``could''.

In most cases, the success rate is reduced by 20\%. However, in the case of UniAct model on LIBERO-Object tasks, the quality dropped by half. 

The LLARP model appears to be fairly robust with human paraphrases, probably due to training on more complex and variant commands.
We also conducted experiments with paraphrases by the DeepSeek V3 model and a template similar to the one used for the crowdsourcing platform. The greatest difference compared to human paraphrases amounts to 14\% and is observed for the LLARP model, which addresses tasks involving navigation. In this case, human paraphrases exhibit greater variability in describing the location and the action to be performed with objects.

\paragraph{Irrelevant Context.}
All models showed performance degradation after adding irrelevant context. 
For a context with the same length as ``Short'', the largest drop in most cases is observed if the noise is semantically and lexically similar to a relevant command from the training set, i.e. belongs to the second group of contexts. 
On these types of contexts, at best a 10\% drop can be observed, but more often models lose more than 50\% of their quality.

An example of how context leads to dysfunctional robot behavior is shown in Fig.~\ref{fig:llarp-scenes}. The target command is specified as `find a lid and move it the black table', while the noise `On the sofa there's an apple' is taken from a set of contexts ``Location''. Pointing to the location of an irrelevant object on the sofa triggers the robot to search for a target object on the sofa. In the absence of an object in the specified location, the robot starts to perform chaotic actions, trying to pick up various non-target objects while moving randomly around the scene.

As the context length increases, the performance of the model starts to decrease consistently for all considered cases. When the context size is equal to the length of the target command, the quality drop for contexts from the first group becomes comparable to the drop on semantically close context types; in some cases, may even surpass it.

\section{Irrelevant Context Filtering}

\subsection{Proposed Framework}
\label{sec:filter}

\begin{figure*}
\begin{tikzpicture}
\begin{axis}[
    width=1.0\textwidth,
    height=6cm,
    ybar,
    enlarge x limits=0.15,
    xlabel={Contexts},
    ylabel={Success rate, \%},
    symbolic x coords={single, short, long, description, infeasible, location},
    xtick=data,
    ymin=0,
    ymax=110,
    ymajorgrids=true,
    grid style=dashed,
    legend style={
        at={(0.5, 1.05)},
        font=\small,
        anchor=south,
        legend columns=6,
        draw=none,
        fill=none,
    },
    every node near coord/.append style={
        font=\tiny,
        rotate=90,
        anchor=west,
        yshift=1pt,
        /pgf/number format/fixed,
        /pgf/number format/precision=1
    },
    ybar legend,
]
\addplot+[ybar, bar width=8.0pt, bar shift=-8.0pt, fill=cyan!70, draw=black, draw opacity=0.6, nodes near coords] coordinates { (single, 96.0000) (short, 89.0000) (long, 88.0000) (description, 94.0600) (infeasible, 76.0000) (location, 39.0000) };
\addplot[ybar, bar width=8.0pt, bar shift=-8.0pt, pattern=north west lines, pattern color=black!70] coordinates { (single, 92.0000) (short, 88.0000) (long, 73.0000) (description, 87.1300) (infeasible, 72.8200) (location, 58.0000) };

\addplot+[ybar, bar width=8.0pt, bar shift=0.0pt, fill=orange!70, draw=black, draw opacity=0.6, nodes near coords] coordinates { (single, 95.0000) (short, 93.0000) (long, 92.0800) (description, 96.0000) (infeasible, 73.2700) (location, 70.0000) };
\addplot[ybar, bar width=8.0pt, bar shift=0.0pt, pattern=crosshatch, pattern color=black!70] coordinates { (single, 83.0000) (short, 80.0000) (long, 80.0000) (description, 76.4700) (infeasible, 48.5100) (location, 55.0000) };

\addplot+[ybar, bar width=8.0pt, bar shift=8.0pt, fill=violet!70, draw=black, draw opacity=0.6, nodes near coords] coordinates { (single, 93.1400) (short, 95.0000) (long, 94.0600) (description, 97.0300) (infeasible, 69.0000) (location, 79.0000) };
\addplot[ybar, bar width=8.0pt, bar shift=8.0pt, pattern=grid, pattern color=black!70] coordinates { (single, 92.0000) (short, 90.0000) (long, 91.1800) (description, 89.1100) (infeasible, 67.0000) (location, 79.0000) };

    \draw[color=red, dotted, thick]
        ({rel axis cs:0,0} |- {axis cs:single,98.3}) -- ({rel axis cs:1,0} |- {axis cs:location,98.3})
        node[pos=0.9, above, font=\small] {Original};

\legend{Flan-t5-base, Qwen2.5-0.5B, Qwen2.5-1.5B, Llama3.2-1B, Qwen2.5-3B, Llama3.2-3B}
\end{axis}
\end{tikzpicture}

\caption{Success rates for LLARP in the Habitat 2.0 simulator for commands with different types of irrelevant context after filtering by LLMs of various sizes using a few-shot prompt.}
\label{fig:filter-llarp-diff-llms}

\end{figure*}
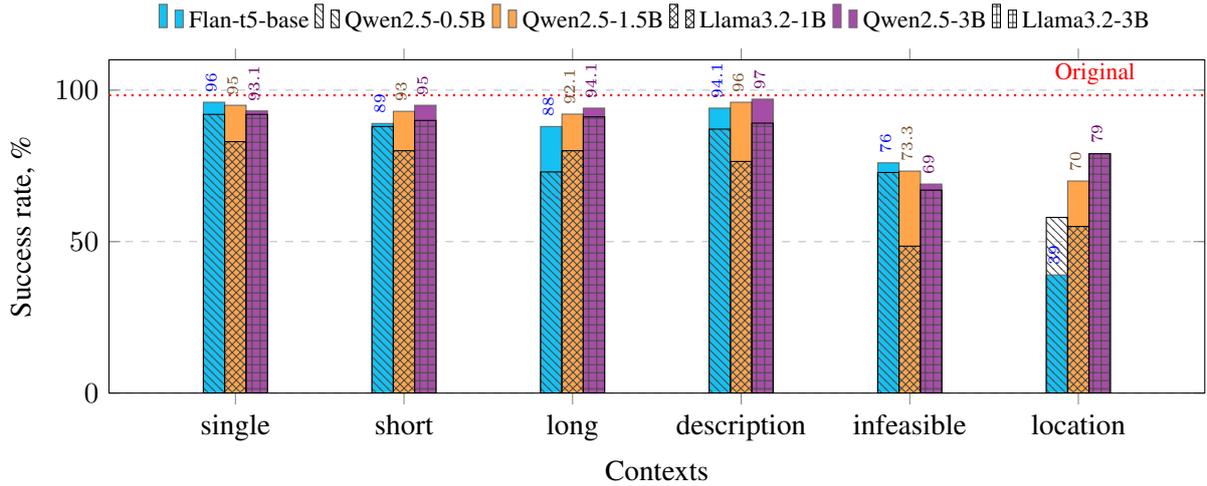

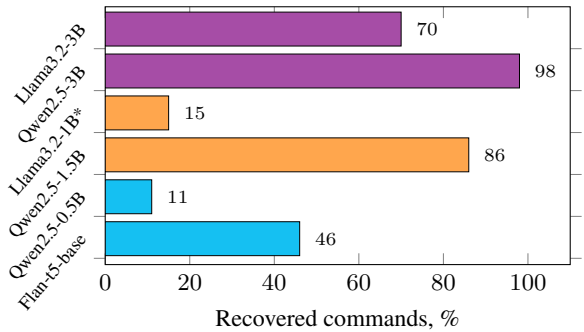
\begin{figure}[t]
\centering
\begin{tikzpicture}
\begin{axis}[
    xbar,
    xlabel={Recovered commands, \%},
    ylabel={},
    ytick={0,1,2,3,4,5},
    yticklabels={Flan-t5-base, Qwen2.5-0.5B, Qwen2.5-1.5B, Llama3.2-1B*, Qwen2.5-3B, Llama3.2-3B},
    yticklabel style={font=\scriptsize, anchor=base east, yshift=10pt, rotate=50},
    xticklabel style={font=\small},
    xlabel style={font=\small},
    xmin=0,
    xmax=110,
    ymin=-0.15,
    ymax=6,
    bar width=0.45cm,
    height=5cm,
    width=\columnwidth,
    nodes near coords,
    nodes near coords style={font=\scriptsize, anchor=west},
    every node near coord/.append style={xshift=2pt},
]
\addplot[fill=cyan!70, draw=black, line width=0.3pt, forget plot] coordinates {(46,0)};     
\addplot[fill=cyan!70, draw=black, line width=0.3pt, forget plot] coordinates {(11,1)};     
\addplot[fill=orange!70, draw=black, line width=0.3pt, forget plot] coordinates {(86,2)};   
\addplot[fill=orange!70, draw=black, line width=0.3pt, forget plot] coordinates {(15,3)};   
\addplot[fill=violet!70, draw=black, line width=0.3pt, forget plot] coordinates {(98,4)};   
\addplot[fill=violet!70, draw=black, line width=0.3pt, forget plot] coordinates {(70,5)};   
\end{axis}
\end{tikzpicture}
\caption{Ratio of recovered commands from the LIBERO benchmark averaged across task suites and all types of irrelevant context}
\label{fig:filter-libero-diff-llm}
\end{figure}

Sec.~\ref{sec:experiments} shows that the presence of irrelevant context leads to undesirable robot behavior. It is essential to extract the main command from the noisy text.
Retraining the VLA model is a computationally and data-intensive process, which does not guarantee improved robustness of the resulting model. Since we consider different types of context, including a complex type in terms of semantic and lexical similarity, it can hardly be processed with templates. Therefore, we address this problem with LLMs, which have been recognized as powerful tools for selective classification, even in zero-shot settings \cite{jeong2025llmselect, tabatabaei-etal-2025-large}. 

We investigate how models of varying sizes—tiny (Flan‑T5 Base, Qwen 2.5 0.5B Instruct), small (Qwen 2.5 1.5B Instruct, Llama 3.2 1B Instruct), medium (Qwen 2.5 3B Instruct, Llama 3.2 3B Instruct), and standard (Meta‑Llama‑3‑8B‑Instruct)—perform on a filtering task in a few-shot setting.
We prompt models with the instruction, which contains three examples of context filtering. Different types of context are used, namely \textit{``Short''}, \textit{``Location''} and \textit{``Infeasible''}.
This prompt is specific and can improve filtering in more complex cases of irrelevant context.
We also examined the instruction with only one context type \textit{``Short''} in the examples. However, it performed poorly on semantically similar contexts (see Table~\ref{tab:app:llarp-irr-filters} in Appendix).

Filtering instructions were adapted for the LLARP model and models for LIBERO benchmarks (see examples in Appendix Figure~\ref{fig:app:prompts-3}).

\subsection{Evaluation of Filtering Framework}
\label{sec:experiments_filtering}

\begin{table*}[t]
\scalebox{0.93}{
\fontsize{11pt}{13pt}\selectfont
\tabcolsep=3pt
\begin{tabular}{cccccccccc}
\toprule
\textbf{Environment}                                                      & \textbf{Model} & \textbf{Original} & \textbf{Single} & \textbf{Short} & \textbf{Long} & \textbf{Location} & \textbf{Description} & \textbf{Infeasible} & \textbf{Human} \\ \hline
\multirow{2}{*}{\begin{tabular}[c]{@{}c@{}}LIBERO\\ Goal\end{tabular}}    
& OpenVLA + F    & 77.5              & 77.5            & 77.5           & 77.5          & 77.5              & 77.5                 & 73.0$\,\downarrow$     & 59.2$\,\uparrow$           \\
                                                                          & UniAct + F     & 67.5              & 67.5            & 67.5           & 67.5          & 67.5              & 67.5                 & 66.0$\,\downarrow$ & 44.2$\,\uparrow$               \\ \hline
\multirow{2}{*}{\begin{tabular}[c]{@{}c@{}}LIBERO\\ Object\end{tabular}}  
& OpenVLA + F    & 87.3              & 87.3            & 87.3           & 87.3          & 87.3              & 87.3                   & 87.3     & 79.1$\,\downarrow$            \\
                                                                          & UniAct + F     & 86.5              & 86.5            & 86.5           & 86.5          & 86.5              & 86.5                 & 86.5    & 45.6$\,\uparrow$            \\ \hline
\multirow{2}{*}{\begin{tabular}[c]{@{}c@{}}LIBERO\\ Spatial\end{tabular}} 
& OpenVLA + F    & 85.3              & 85.3            & 85.3           & 85.3          & 85.3              & 85.3                 & 85.3      & 55.0$\,\downarrow$          \\
                                                                          & UniAct + F     & 79.0              & 79.0            & 79.0           & 79.0          & 79.0              & 79.0                 & 79.0          & 48.0$\,\downarrow$      \\ \hline
\multirow{3}{*}{\begin{tabular}[c]{@{}c@{}}LIBERO\\ Long\end{tabular}}    
& OpenVLA + F    & 51.0$\,\downarrow$             & 51.7            & 51.7           & 51.7          & 51.7              & 51.7                 & 46.7$\,\downarrow$      & 35.0$\,\downarrow$          \\
                                                                          & UniAct + F     & 46.5              & 46.5            & 46.5           & 46.5          & 46.5              & 46.5                 & 37.5$\,\downarrow$ & 19.6$\,\uparrow$               \\
                                                                          & MoDE + F       & 95.5              & 95.5            & 95.5           & 95.5          & 95.5              & 95.5                 & 93.5$\,\downarrow$  & -             \\ \hline
Habitat 2.0                                                               
& LLARP + F      & 98.3              & 98.3            & 98.3           & 98.3          & 98.3              & 95.7$\,\downarrow$                & 94.9$\,\downarrow$  & 82.1$\,\downarrow$             \\ \bottomrule
\end{tabular}
  }
  \label{tab:filter}
  \caption{
    Success rates of models on the LIBERO benchmark and Habitat 2.0 simulator on commands after filtering with Meta‑Llama‑3‑8B‑Instruct. Arrows correspond to cases where original commands were not fully recovered.
  }
  \label{tab:filtering}
\end{table*}

The filter behaves differently on noisy commands for the LIBERO benchmark versus the LLARP model, due to differences in the underlying target commands. While LIBERO uses template-style commands (Appendix Tab.~\ref{tab:context_examples:3}), LLARP was trained on more natural language with the navigation part (Appendix Tab.~\ref{tab:context_examples:1}). As a result, on noisy LIBERO commands the filter does not change the target command regardless of whether it succeeds in detecting the context or not. For LLARP, removing context can lead to paraphrasing. 

Figure~\ref{fig:filter-llarp-diff-llms} illustrates how the number of successes varies for LLARP in the Habitat 2.0 simulator depending on filtering by LLMs with sizes up to 3B. As can be seen from the figure, even small filters with up to 0.5B parameters handle filtering of random context well. However, as semantic similarity increases, the quality of filtering decreases and becomes comparable to the results before filtering for Flat-T5 Base for the context type \textit{``Location''}, while Llama 3.2 3B Instruct demonstrates the maximum gain up to 79\% of successful episodes, which is still lower then the original quality.

For LIBERO commands, starting at the 3B model size, only Llama 3.2 is able to recover the majority of the original commands (see Figure~\ref{fig:filter-libero-diff-llm}). It should be noted, that Llama 3.2 1B Instruct had difficulty following the template in the instruction, and its results underwent minor post-processing, where the filtered command was extracted from the overall generated text that contained variants of the phrase ‘filtered:’, ‘filter:’, and etc.

If we further increase the model size and examine the Meta-Llama-3-8B-Instruct model, the detailed analysis shows the following. 
In the case of the LIBERO template, almost all types of irrelevant context were filtered out successfully (see Table~\ref{tab:filtering}), and the target command remained unchanged. 
The only exceptions were commands that were preceded by infeasible non-target commands of the type \textit{``Infeasible''}. For the LLARP model and VLA models on LIBERO-Goal and LIBERO-Long benchmarks, the performance is recovered by more than 90\%.






\subsection{Discussion}
\label{sec:discussion}









Processing of the original target commands and human paraphrased commands with the proposed filtering framework revealed certain issues. 
When applied to human paraphrases, the filter accidentally removed potentially useful information from 5\% of the commands. Therefore, the number of successes could decrease by a few percent, but in some cases, we observe a quality improvement due to the filtering of irrelevant words and the standardization of commands (LIBERO Object and Long suits in Tab.~\ref{tab:filtering}).
The commands for LLARP model could originally go in a more complex form, since the original text instructions contained information about the location of the object. When filtering, these additional details could be classified as irrelevant context and filtered out along with the noise (see Table~\ref{tab:filtering-incorrect} in Appendix).
This causes the quality on the \textit{`Description'} context set to not fully recovered and left a 4\% gap despite full noise filtering (Tab.~\ref{tab:filtering}).
A similar situation was observed for one original language command from the LIBERO-Long benchmark (see Table~\ref{tab:filtering-incorrect} in Appendix). This caused a 0.7\% decrease in initial quality (Tab.~\ref{tab:filtering}).
Despite these cases, the filtering framework significantly improved overall performance. The described incidents were rare and occurred in 0.6\% of our test data.

\section{Conclusion}
\label{sec:conclusions}

This study has thoroughly investigated the vulnerability of current vision-language-action models to human paraphrases  and the presence of irrelevant linguistic context in robot manipulation commands. 
Experiments have shown that even minor textual noise can drastically reduce task success rates, with models showing pronounced sensitivity to certain types of irrelevant context. This behavior generalizes across VLA models based on different LLMs and is observed across various benchmarks and simulators. Employing LLMs as filters to preprocess and clean noisy instructions proves effective in enhancing robustness and restoring performance. 
Evaluating human-generated paraphrases further underscores the current limitations in the robustness of VLA models, which have primarily been trained and tested using synthetic data.
Future research could focus on providing adaptive filtering, processing more complex commands, and improving the robustness of robot understanding for reliable real-world deployment. 
Overall, this work highlights the critical importance of addressing linguistic variability to develop practical and widely utilized embodied AI systems.


\section*{Limitations}
We have considered several reasonable groups of irrelevant context, but  leave aside target commands with conditions and reasoning tasks, as these have been separately investigated in other works. The proposed filtering method, while helping to restore the quality of the model in general, may occasionally filter out some important details. However, this is a rare occurrence and it affected about 1\% of the language commands examined in this paper.

\section*{Ethics}

Our work introduces a novel irrelevant context generation method to evaluate its impact on VLA robotic models. We acknowledge that our method for generating irrelevant linguistic context might be exploited to deliberately confuse deployed VLA systems. Nevertheless, we are convinced that the scientific value of openly documenting these vulnerabilities outweighs that misuse risk. By shedding light on VLA models' failures, we aim to catalyze safer and reliable embodied agents, and will release all code and data under a research-only license to promote responsible use.

Our study involves using crowdsourcing with paid participants to collect paraphrases of embodied AI commands created by humans. We paid crowdworkers at rates above the average wage to ensure fair compensation for their time and effort. This approach reflects our commitment to work ethics and respects the value of human contributions to AI research. 

\paragraph{Crowdsourcing} We used Toloka.ai as a crowdsourcing vendor. According to the user agreement and privacy policy, personal data typically includes information that can identify an individual, such as name, contact information, and other personal identifiers. Human paraphrases do not fall under this category. Moreover, we provide fully anonymized data that can not be linked to the people who wrote each text. Toloka policy allows for the sharing of anonymized data with third parties.



\newpage
\clearpage

\appendix

\section{Appendix}
\label{sec:appendix}

In order to complete all the evaluations, we spent 1300 GPU hours utilizing 5 NVIDIA Tesla
A100 GPUs.

\subsection{ Examples of Commands with Irrelevant Context and results}
\label{sec:app:ex-ir}

This subsection provides concrete examples of noisy commands used to evaluate the impact of irrelevant context on VLA models across different simulation environments (Habitat 2.0 and LIBERO benchmarks). The commands in Table~\ref{tab:context_examples:1}--~\ref{tab:context_examples:4} illustrate the insertion of various types of irrelevant context around the original robot commands to test model robustness.

These tables highlight the diversity and complexity of noise introduced to test model vulnerability.

Table~\ref{tab:app:openvla-irr-filters} and~\ref{tab:app:llarp-irr-filters} show the differences in the effect of noise inserted before and after the command.


\begin{table*}[h!]
\centering
\renewcommand{\arraystretch}{1.3}
\begin{tabular}{l|p{14cm}}
\hline
\textbf{Type} & \textbf{Command} \\
\hline
Original & Find a sponge and move it to the right counter.

Find a can and move it to the brown table.

Find a strawberry and move it to the black table.

Find a screwdriver and move it to the right counter.

On the sink I need you to put a lego.

Find a lemon and move it to the sofa.

Find a bowl and move it to the right counter.

Find a can and move it to the left counter.

On the right counter I need you to put a knife.

On the sink I need you to put a ball. \\
\hline
Single   & 
Although, find a mug on the right counter and move it to the tv stand.

Meanwhile, find a wrench and move it to the left counter.

Furthermore, find a orange and move it to the right counter.

Consequently, find a lid and move it to the black table.

Additionally, find a lid on the left counter and move it to the black table.

Thus, find a wrench on the right counter and move it to the left counter.

Nevertheless, find a mug and move it to the brown table.

Regardless, find a wrench on the right counter and move it to the sink.

Otherwise, find a mug and move it to the sofa.

Whereas, find a toy airplane and move it to the right counter. \\
\hline
Short    & Joyful moment watching birds. Find a orange and move it to the brown table.

Anxiety peaks near deadlines. Find a toy airplane and move it to the right counter.

Laughing while watching movie. Find a orange and move it to the brown table.

Relieved after finishing chores. Find a toy airplane on the sink and move it to the black table.

Excited gaming with friends. Find a wrench on the brown table and move it to the right counter.

Embarrassed dancing in pajamas. Find a orange and move it to the right counter.

Find a wrench and move it to the right counter. Inspired while painting walls.

Regretful about skipped cleaning. Find a toy airplane and move it to the black table.

Happy singing in shower. Find a lid on the black table and move it to the right counter.

Find a lid and move it to the left counter. Guilty for eating snacks.  \\
\hline
Long     & He felt motivated cleaning the pantry and organizing everything, so find an orange and move it to the sink.

He felt nostalgic holding his childhood paint once again. Find a mug on the right counter and move it to the TV stand.

A burst of joy came with finishing her knitting project. Find a wrench and move it to the left counter.

Late at night, anxiety crept in with pending homework reminders. Find a orange and move it to the right counter.

She felt silly dancing barefoot to her favorite songs. Find a lid and move it to the black table.

Inspiration struck while staring at the blank white canvas. Find a lid on the left counter and move it to the black table.

Find a wrench on the right counter and move it to the left counter. He was satisfied after prepping meals for the entire week.

Find a mug and move it to the brown table. The gloomy weather matched her tired and melancholy mood today.

Find a wrench on the right counter and move it to the sink. She nervously reorganized shelves to distract from her future worries.

Find a mug and move it to the sofa. Music played, bringing peace to once-chaotic morning. \\
\hline
\end{tabular}
\caption{Examples of context inserted into commands for the LLARP model in the Habitat 2.0 simulator.}
\label{tab:context_examples:1}
\end{table*}

\begin{table*}[h!]
\centering
\renewcommand{\arraystretch}{1.3}
\begin{tabular}{l|p{14cm}}
\hline
\textbf{Type} & \textbf{Command} \\
\hline
Location & There’s an apple on the TV stand, but find an orange and move it to the sink.

On the sofa there's a lemon, but find a mug on the right counter and move it to the TV stand.

On the sofa there's an apple, but instead find a wrench and move it to the left counter.

Find a orange and move it to the right counter. I left my lid on the black table.

There’s a wrench in the sink, but find a lid and move it to the black table.

On the brown table there's an lego, but instead find a lid on the left counter and move it to the black table. I left my toy airplane on the brown table.

Find a wrench on the right counter and move it to the left counter. I left my lid on the black table. 

Find a mug and move it to the brown table. There’s an orange on the right counter.

Find a wrench on the right counter and move it to the sink. There’s a lego on the sofa.

Find a plum and move it to the sofa. I left my apple on the left counter. 
\\
\hline
Description & Cup is a container for liquids. Find an orange and move it to the sink.

Clamp is useful for woodworking. Find a mug on the right counter and move it to the TV stand.

Lemon is citrus. Find a lid and move it to the sink.

Orange is juicy fruit. Find a wrench on the left counter and move it to the right counter.

Plum is a fruit. Find a mug and move it to the sofa.

Lego has colorful blocks. Bring the wrench from the right counter to the left counter.

Bring the bowl from the TV stand to the right counter? Lid protects from spilling.

Find a toy airplane on the right counter and move it to the left counter. Strawberry is delicious.

Find a mug and move it to the sink. Bowl is for serving food.
Find a mug and move it to the black table. Ball is a round object.\\
\hline
Infeasible  & Bake a pie with peach slices. Find an orange and move it to the sink. 

Cut ribbons with sharp scissors. Find a mug and move it to the sofa.

Spray cleanser on the countertops. Bring the toy airplane from the right counter to the left counter.

Bite into the soft plum slowly. Find a mug and move it to the sofa.

Garnish tea with fresh lemon slices. Find a lid and move it to the sink.

Bounce a small ball around. Bring the wrench from the sofa to the TV stand.

Clamp the pipe for repair. Find a lid and move it to the left counter.

Find a orange and move it to the right counter. Kick the colorful ball outside.

Bite into the soft plum slowly. Glide the toy airplane across the room.

Bring the wrench from the sofa to the TV stand. Explore new stories in your book. \\
\hline
\end{tabular}
\caption{Examples of context inserted into commands for the LLARP model in the Habitat 2.0 simulator.}
\label{tab:context_examples:2}
\end{table*}

\begin{table*}[h!]
\centering
\renewcommand{\arraystretch}{1.3}
\begin{tabular}{l|p{14cm}}
\hline
\textbf{Type} & \textbf{Command} \\
\hline
Original & turn on the stove

put the bowl on top of the cabinet

put the wine bottle on the rack

pick up the cream cheese and place it in the basket

pick up the milk and place it in the basket

pick up the chocolate pudding and place it in the basket

pick up the black bowl between the plate and the ramekin and place it on the plate

pick up the black bowl next to the ramekin and place it on the plate

put both the cream cheese box and the butter in the basket

put both moka pots on the stove
\\
\hline
Single   & however open the middle drawer of the cabinet

therefore put the bowl on the stove
moreover put the wine bottle on top of the cabinet

open the top drawer and put the bowl inside although

meanwhile put the bowl on top of the cabinet

furthermore push the plate to the front of the stove

consequently put the cream cheese in the bowl

turn on the stove
thus put the bowl on the plate additionally

nevertheless put the wine bottle on the rack \\
\hline
Short    & open the middle drawer of the cabinet feeling lonely in bed

morning coffee feels serene put the bowl on the stove

restless pacing after sunset put the wine bottle on top of the cabinet

nostalgia strikes after dinner open the top drawer and put the bowl inside

put the bowl on top of the cabinet confused thoughts while cooking

push the plate to the front of the stove quiet satisfaction reading books

focusing deeply at night put the cream cheese in the bowl

stress builds during arguments turn on the stove

calm mornings with tea put the bowl on the plate

peaceful napping on weekends put the wine bottle on the rack \\
\hline
Long     & she found peace sipping tea in the afternoon light open the middle drawer of the cabinet

frustration loomed as he searched for his misplaced keys put the bowl on the stove

quiet happiness filled the room while rearranging photo albums put the wine bottle on top of the cabinet

he felt nostalgic holding his childhood toy once again open the top drawer and put the bowl inside

a burst of joy came with finishing her knitting project put the bowl on top of the cabinet

push the plate to the front of the stove late at night anxiety crept in with pending homework reminders

she felt silly dancing barefoot to her favorite songs put the cream cheese in the bowl

turn on the stove inspiration struck while staring at the blank white canvas

he was satisfied after prepping meals for the entire week put the bowl on the plate

the gloomy weather matched her tired and melancholy mood today put the wine bottle on the rack \\
\hline
\end{tabular}
\caption{Examples of context inserted into commands for models evaluated on the LIBERO benchmark.}
\label{tab:context_examples:3}
\end{table*}

\begin{table*}[h!]
\centering
\renewcommand{\arraystretch}{1.3}
\begin{tabular}{l|p{14cm}}
\hline
\textbf{Type} & \textbf{Command} \\
Location & open the middle drawer of the cabinet the bowl is in the stove

the mug is in the microwave put the bowl on the stove

put the wine bottle on top of the cabinet the moka pots are on the stove

the cream cheese is in the basket open the top drawer and put the bowl inside

the mug is on the plate put the bowl on top of the cabinet

 push the plate to the front of the stove the chocolate pudding is on the plate

put the cream cheese in the bowl the mug is on the right plate

turn on the stove the book is inside of the caddy

put the bowl on the plate the wine bottle is in the basket

the wine bottle in the cabinet pick up the milk and place it in the basket \\
\hline
Description & open the middle drawer of the cabinet ball is a round object

put the bowl on the stove ball bounces on surfaces

clamp holds objects together put the wine bottle on top of the cabinet

clamp is useful for woodworking open the top drawer and put the bowl inside

hammer is tool for driving nails put the bowl on top of the cabinet

push the plate to the front of the stove hammer has a metal head

put the cream cheese in the bowl screwdriver tightens screws

screwdriver comes in various sizes turn on the stove

put the bowl on the plate padlock secures items with a key

padlock are made of metal put the wine bottle on the rack \\
\hline
Infeasible  & toss the ball softly outdoors put both the alphabet soup and the tomato sauce in the basket

put both the cream cheese box and the butter in the basket enjoy a juicy plum after lunch

snack on a sweet pear tonight turn on the stove and put the moka pot on it

a ripe peach feels refreshing put the black bowl in the bottom drawer of the cabinet and close it

put the white mug on the left plate and put the yellow and white mug on the right plate bite into a crisp apple

pick up the book and place it in the back compartment of the caddy eat fresh strawberries with yogurt

solve the tricky rubik’s cube put the white mug on the plate and put the chocolate pudding to the right of the plate

read your favorite book tonight put both the alphabet soup and the cream cheese box in the basket

put both moka pots on the stove sip tea from the cup quietly

fly the toy airplane for fun put the yellow and white mug in the microwave and close it
\\
\hline
\end{tabular}
\caption{Examples of context inserted into commands for models evaluated on the LIBERO benchmark.}
\label{tab:context_examples:4}
\end{table*}

\subsection{Instructions examples and results of the filtering framework}
\label{sec:app:filter}

This subsection presents the prompt instructions for the first and second types filtering for both Habitat 2.0 (LLARP model) and the LIBERO benchmark.

Each prompt from Figure~\ref{fig:app:prompts-3} includes three examples of filtering out short, location and infeasible types of irrelevant phrases that do not refer to scene objects or commands.

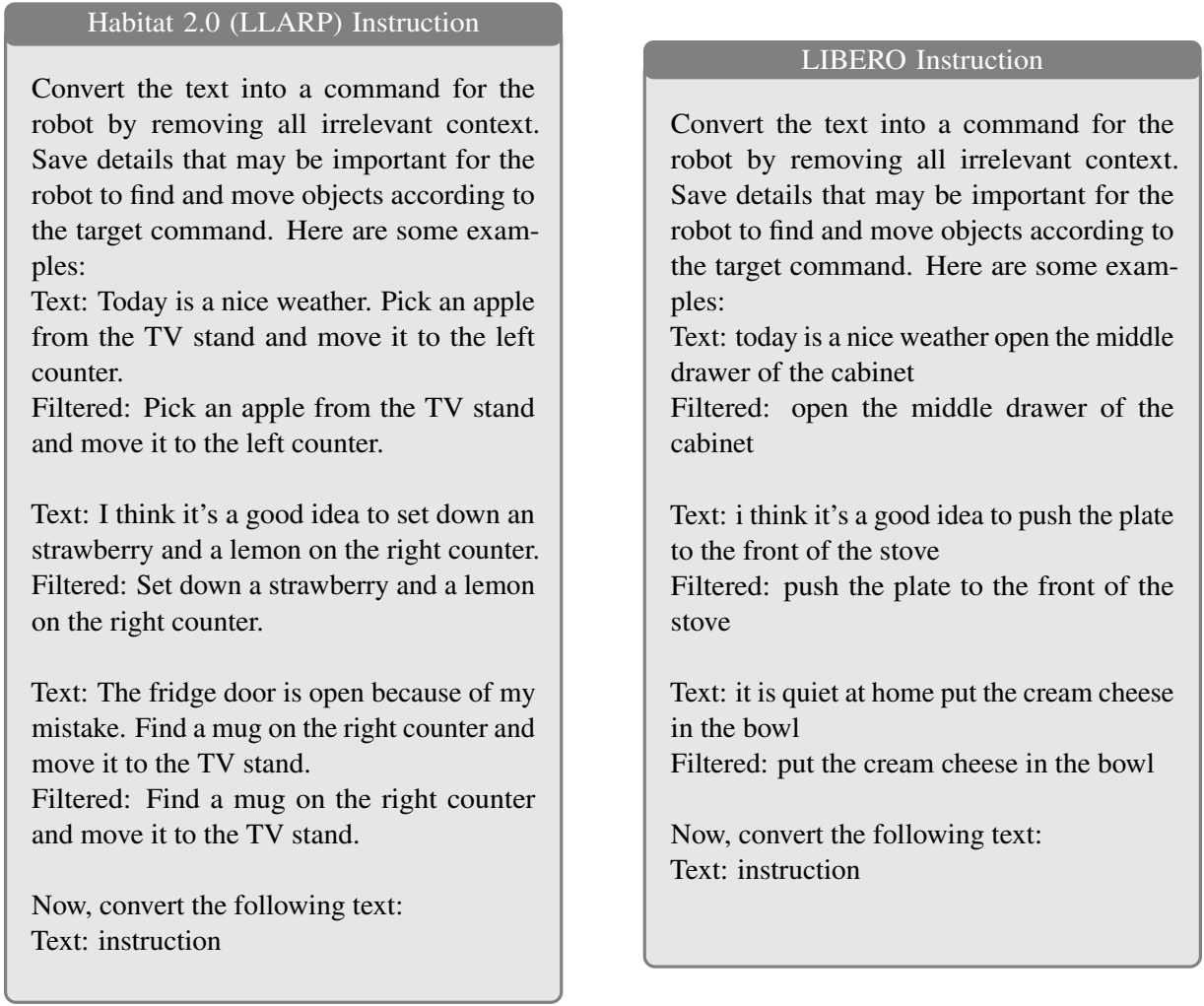
\begin{figure*}[t]
  \tikzstyle{mybox} = [draw=gray, fill=gray!20, very thick,
    rectangle, rounded corners, inner sep=10pt, inner ysep=20pt]
  \tikzstyle{fancytitle} = [fill=gray, text=white]
  
  \begin{tikzpicture}[baseline=-2cm]
    \node [mybox] (box){
      \begin{minipage}[t!]{0.42\textwidth}
    Convert the text into a command for the robot by removing all irrelevant context. Save details that may be important for the robot to find and move objects according to the target command. Here are some examples:
   
    Text: Today is a nice weather. Pick an apple from the TV stand and move it to the left counter.
    
    Filtered: Pick an apple from the TV stand and move it to the left counter.
    \hfill \break

    Text: I think it’s a good idea to set down an strawberry and a lemon on the right counter.
    
    Filtered: Set down a strawberry and a lemon on the right counter.
    \hfill \break

    Text: The fridge door is open because of my mistake. Find a mug on the right counter and move it to the TV stand.
    
    Filtered: Find a mug on the right counter and move it to the TV stand.
    \hfill \break

    Now, convert the following text:
    
    Text: {instruction}
      \end{minipage}
    };
    \node[fancytitle, text width=0.45\textwidth, text centered, rounded corners] at (box.north) {Habitat 2.0 (LLARP) Instruction};
  \end{tikzpicture}
\hfill
    \begin{tikzpicture}[baseline=-2cm]
    \node [mybox] (box){
      \begin{minipage}[t!]{0.42\textwidth}
    Convert the text into a command for the robot by removing all irrelevant context. Save details that may be important for the robot to find and move objects according to the target command. Here are some examples:
    
    Text: today is a nice weather open the middle drawer of the cabinet
    
    Filtered: open the middle drawer of the cabinet
    \hfill \break

    Text: i think it’s a good idea to push the plate to the front of the stove
    
    Filtered: push the plate to the front of the stove
    \hfill \break

    Text: it is quiet at home put the cream cheese in the bowl
    
    Filtered: put the cream cheese in the bowl
    \hfill \break

    Now, convert the following text:
    
    Text: {instruction}
    \hfill \break
    
      \end{minipage}
    };
    \node[fancytitle, text width=0.45\textwidth, text centered, rounded corners] at (box.north) {LIBERO Instruction};
  \end{tikzpicture}
  \caption{Examples of instructions with 1 type of irrelevant context used in the filtering framework.}
  \label{fig:app:prompts}
\end{figure*}

Each prompt from Figure~\ref{fig:app:prompts} includes three examples of filtering out short irrelevant phrases that do not refer to scene objects or commands, i.e. context of the type \textit{``Short''}. 
This type of context does not contain information about the training data. It allows to assess how generalizable a given filtering method is to other types of context. However, we found that this type of prompt demonstrates poor performance when filtering semantically similar contexts (Table~\ref{tab:app:llarp-irr-filters}), therefore all results in the main sections are presented for the second type of instruction.

\begin{figure*}[t]
  \tikzstyle{mybox} = [draw=gray, fill=gray!20, very thick,
    rectangle, rounded corners, inner sep=10pt, inner ysep=20pt]
  \tikzstyle{fancytitle} = [fill=gray, text=white]
  
  \begin{tikzpicture}[baseline=-2cm]
    \node [mybox] (box){
      \begin{minipage}[t!]{0.42\textwidth}
    Convert the text into a command for the robot by removing all irrelevant context. Save details that may be important for the robot to find and move objects according to the target command. Here are some examples:
   
    Text: Today is a nice weather. Pick an apple from the TV stand and move it to the left counter.
    
    Filtered: Pick an apple from the TV stand and move it to the left counter.
    \hfill \break

    Text: There's an apple in the fridge, but set down a strawberry and a lemon on the right counter.
    
    Filtered: Set down a strawberry and a lemon on the right counter.
    \hfill \break

    Text: Tighten screws with a screwdriver. Find a mug on the right counter and move it to the TV stand.
    
    Filtered: Find a mug on the right counter and move it to the TV stand.
    \hfill \break

    Now, convert the following text:
    
    Text: {instruction}
      \end{minipage}
    };
    \node[fancytitle, text width=0.45\textwidth, text centered, rounded corners] at (box.north) {Habitat 2.0 (LLARP) Instruction};
  \end{tikzpicture}
\hfill
    \begin{tikzpicture}[baseline=-2cm]
    \node [mybox] (box){
      \begin{minipage}[t!]{0.42\textwidth}
    Convert the text into a command for the robot by removing all irrelevant context. Save details that may be important for the robot to find and move objects according to the target command. Here are some examples:
    
    Text: today is a nice weather open the middle drawer of the cabinet
    
    Filtered: open the middle drawer of the cabinet
    \hfill \break

    Text: the mug is on the right plate pick up the black bowl between the plate and the ramekin and place it on the plate
    
    Filtered: pick up the black bowl between the plate and the ramekin and place it on the plate
    \hfill \break

    Text: spray the counter with cleanser pick up the cream cheese and place it in the bowl
    
    Filtered: pick up the cream cheese and place it in the bowl
    \hfill \break

    Now, convert the following text:
    
    Text: {instruction}
    
      \end{minipage}
    };
    \node[fancytitle, text width=0.45\textwidth, text centered, rounded corners] at (box.north) {LIBERO Instruction};
  \end{tikzpicture}
  \caption{Examples of instructions with 3 different types of irrelevant context used in the filtering framework.}
  \label{fig:app:prompts-3}
\end{figure*}

This approach relies on few-shot prompting with LLMs, demonstrating its ability to discard irrelevant context effectively without knowledge of the robot's training process or task feasibility.

Table~\ref{tab:app:llarp-irr-filters} demonstrates how this type of prompt instructions can generalize filtering across different types of context. 
As can be seen from the table, the generalization is generally present, but ``Infeasible'' type of noise requires additional information or examples about the robot's abilities.

Table~\ref{tab:filtering-incorrect} highlights the potential pitfalls for filtering framework, when important details can be accidentally removed.

\begin{table*}[t]
\centering
\tabcolsep=4.5pt
\begin{tabular}{ccccccccc}
\toprule
\multicolumn{2}{c}{\textbf{Setup}} & \textbf{Original} & \textbf{Linking} & \textbf{Short} & \textbf{Long} & \textbf{Location} & \textbf{Description} & \textbf{Infeasible} \\ \hline
\multirow{2}{*}{Goal} &
Noise Before &
\gradientnums{20.5}{77.5}{77.5} &
\gradientnums{20.5}{77.5}{71.5} &
\gradientnums{20.5}{77.5}{48.0} &
\gradientnums{20.5}{77.5}{20.5} &
\gradientnums{20.5}{77.5}{29.0} &
\gradientnums{20.5}{77.5}{35.5} &
\gradientnums{20.5}{77.5}{29.0} \\

& Noise After &
\gradientnums{17.3}{77.5}{77.5} &
\gradientnums{17.3}{77.5}{63.7} &
\gradientnums{17.3}{77.5}{39.0} &
\gradientnums{17.3}{77.5}{17.3} &
\gradientnums{17.3}{77.5}{22.0} &
\gradientnums{17.3}{77.5}{25.3} &
\gradientnums{17.3}{77.5}{27.0} \\
\hline

\multirow{2}{*}{Object} &
Noise Before &
\gradientnums{53.0}{87.3}{87.3} &
\gradientnums{53.0}{87.3}{86.3} &
\gradientnums{53.0}{87.3}{77.7} &
\gradientnums{53.0}{87.3}{53.0} &
\gradientnums{53.0}{87.3}{74.0} &
\gradientnums{53.0}{87.3}{72.3} &
\gradientnums{53.0}{87.3}{64.0} \\

& Noise After &
\gradientnums{59.3}{87.3}{87.3} &
\gradientnums{59.3}{87.3}{86.3} &
\gradientnums{59.3}{87.3}{70.7} &
\gradientnums{59.3}{87.3}{59.3} &
\gradientnums{59.3}{87.3}{71.0} &
\gradientnums{59.3}{87.3}{68.3} &
\gradientnums{59.3}{87.3}{61.0} \\
\hline

\multirow{2}{*}{Spatial} &
Noise Before &
\gradientnums{56.0}{85.3}{85.3} &
\gradientnums{56.0}{85.3}{84.3} &
\gradientnums{56.0}{85.3}{75.0} &
\gradientnums{56.0}{85.3}{56.0} &
\gradientnums{56.0}{85.3}{67.0} &
\gradientnums{56.0}{85.3}{73.7} &
\gradientnums{56.0}{85.3}{64.0} \\

& Noise After &
\gradientnums{48.0}{85.3}{85.3} &
\gradientnums{48.0}{85.3}{79.7} &
\gradientnums{48.0}{85.3}{58.0} &
\gradientnums{48.0}{85.3}{48.0} &
\gradientnums{48.0}{85.3}{58.0} &
\gradientnums{48.0}{85.3}{50.0} &
\gradientnums{48.0}{85.3}{59.0} \\
\hline

\multirow{2}{*}{Long} &
Noise Before &
\gradientnums{31.7}{51.7}{51.7} &
\gradientnums{31.7}{51.7}{51.3} &
\gradientnums{31.7}{51.7}{35.3} &
\gradientnums{31.7}{51.7}{31.7} &
\gradientnums{31.7}{51.7}{25.0} &
\gradientnums{31.7}{51.7}{40.3} &
\gradientnums{31.7}{51.7}{32.0} \\

& Noise After &
\gradientnums{29.3}{51.7}{51.7} &
\gradientnums{29.3}{51.7}{45.7} &
\gradientnums{29.3}{51.7}{30.3} &
\gradientnums{29.3}{51.7}{29.3} &
\gradientnums{29.3}{51.7}{22.0} &
\gradientnums{29.3}{51.7}{31.7} &
\gradientnums{29.3}{51.7}{29.0} \\
\hline
\end{tabular}
    \caption{\label{tab:app:openvla-irr-filters} Success rate of the OpenVLA model on the LIBERO-Goal, Object, Spatial and Long task suits depending on irrelevant context, color-coded by value magnitude.
  }
\end{table*}

\begin{table*}[t]
\centering
\tabcolsep=4.5pt
\begin{tabular}{ccccccccc}
\toprule
\multicolumn{2}{c}{\textbf{Setup}} & \textbf{Original} & \textbf{Linking} & \textbf{Short} & \textbf{Long} & \textbf{Location} & \textbf{Description} & \textbf{Infeasible} \\ \hline
\multirow{2}{*}{Goal} &
Noise Before &
\gradientnums{47.0}{91.5}{91.5} &
\gradientnums{47.0}{91.0}{91.3} &
\gradientnums{47.0}{91.0}{82.8} &
\gradientnums{47.0}{91.0}{49.8} &
\gradientnums{59.5}{74.0}{60.0} &
\gradientnums{59.5}{74.0}{75.8} &
\gradientnums{59.5}{74.0}{69.0} \\ 

& Noise After &
\gradientnums{17.3}{77.5}{91.5} &
\gradientnums{17.3}{77.5}{92.0} &
\gradientnums{17.3}{77.5}{73.0} &
\gradientnums{17.3}{77.5}{39.8} &
\gradientnums{17.3}{77.5}{51.3} &
\gradientnums{17.3}{77.5}{61.8} &
\gradientnums{17.3}{77.5}{50.3} \\
\hline

\multirow{2}{*}{Object} &
Noise Before &
\gradientnums{83.5}{98}{97.5} &
\gradientnums{83.5}{98}{97.3} &
\gradientnums{83.5}{98}{94.5} &
\gradientnums{83.5}{98}{83.8} &
\gradientnums{89.0}{92}{85.0} &
\gradientnums{89.0}{92}{92.5} &
\gradientnums{89.0}{92}{92.5}\\ 

& Noise After &
\gradientnums{59.3}{87.3}{97.5} &
\gradientnums{59.3}{87.3}{97.5} &
\gradientnums{59.3}{87.3}{95.0} &
\gradientnums{59.3}{87.3}{86.0} &
\gradientnums{59.3}{87.3}{86.8} &
\gradientnums{59.3}{87.3}{91.0} &
\gradientnums{59.3}{87.3}{92.5} \\
\hline

\multirow{2}{*}{Spatial} &
Noise Before &
\gradientnums{82.0}{97}{96.75} &
\gradientnums{82.0}{97}{97.0} &
\gradientnums{82.0}{97}{94.0} &
\gradientnums{82.0}{97}{77.0} &
\gradientnums{82.0}{97}{82.0} &
\gradientnums{82.0}{97}{92.0} &
\gradientnums{82.0}{97}{90.5}\\ 

& Noise After &
\gradientnums{48.0}{85.3}{96.75} &
\gradientnums{48.0}{85.3}{97.0} &
\gradientnums{48.0}{85.3}{94.5} &
\gradientnums{48.0}{85.3}{79.8} &
\gradientnums{48.0}{85.3}{81.5} &
\gradientnums{48.0}{85.3}{93.0} &
\gradientnums{48.0}{85.3}{88.8} \\
\hline

\multirow{2}{*}{Long} &
Noise Before &
\gradientnums{66.5}{88}{88.5} &
\gradientnums{66.5}{88}{85.0} &
\gradientnums{66.5}{88}{79.0} &
\gradientnums{66.5}{88}{66.5} &
\gradientnums{66.5}{88}{72.5} &
\gradientnums{66.5}{88}{80.0} &
\gradientnums{66.5}{88}{84.5}\\

& Noise After &
\gradientnums{29.3}{51.7}{88.5} &
\gradientnums{29.3}{51.7}{85.3} &
\gradientnums{29.3}{51.7}{82.5} &
\gradientnums{29.3}{51.7}{66.0} &
\gradientnums{29.3}{51.7}{74.0} &
\gradientnums{29.3}{51.7}{80.5} &
\gradientnums{29.3}{51.7}{83.8} \\
\hline
\end{tabular}
    \caption{\label{tab:app:openpi-irr-filters} Success rate of the $\pi_0$ model on the LIBERO-Goal, Object, Spatial and Long task suits depending on irrelevant context, color-coded by value magnitude.
  }
\end{table*}

\begin{table*}[t]
\centering
\tabcolsep=4.5pt
\begin{tabular}{cccccccc}
\toprule
\textbf{Setup} & \textbf{Original} & \textbf{Linking} & \textbf{Short} & \textbf{Long} & \textbf{Location} & \textbf{Description} & \textbf{Infeasible} \\ \hline
Noise Before &
\gradientnums{46.2}{98.3}{98.3} &
\gradientnums{46.2}{98.3}{97.5} &
\gradientnums{46.2}{98.3}{90.8} &
\gradientnums{46.2}{98.3}{60.7} &
\gradientnums{46.2}{98.3}{46.2} &
\gradientnums{46.2}{98.3}{89.8} &
\gradientnums{46.2}{98.3}{57.8} \\

Noise After &
\gradientnums{58.9}{98.3}{98.3} &
\gradientnums{58.9}{98.3}{97.3} &
\gradientnums{58.9}{98.3}{93.1} &
\gradientnums{58.9}{98.3}{62.7} &
\gradientnums{58.9}{98.3}{58.9} &
\gradientnums{58.9}{98.3}{75.9} &
\gradientnums{58.9}{98.3}{65.0} \\

Filter 3-shots 1-type  & 98.3 & 98.3 & 98.3 & 98.3 & 98.3 & 96.0 & 77.6 \\ 
Filter 3-shots 3-types & 98.3 & 98.3 & 98.3 & 98.3 & 98.3 & 95.7 & 94.9 \\ \bottomrule
\hline
\end{tabular}
\caption{\label{tab:app:llarp-irr-filters}Success rate of the LLARP model in the Habitat 2.0 simulator depending on irrelevant context, color-coded by value magnitude.}
\end{table*}

\begin{table*}[h!]
\centering
\renewcommand{\arraystretch}{1.3}
\begin{tabular}{p{8cm}|p{8cm}}
\hline
\textbf{Before filtering} & \textbf{After filtering} \\
\hline
Scissors has sharp blades. Find a lid on the right counter and bring it to the sofa.
&
Bring it to the sofa. \\
\hdashline
Spoon scoops liquids. Find the wrench in the sink and bring it to the sofa.
&
Bring the wrench to the sofa. \\
\hdashline
Peach is for desserts. Find the wrench on the sofa and bring it to the TV stand.
&
Bring the wrench to the TV stand. \\
\hdashline
Spatula scrapes food. Find the wrench on the sofa and bring it to the brown table.
&
Bring the wrench to the brown table.
\\
\hdashline
turn on the stove and put the moka pot on it & put the moka pot on it \\
\hline
\end{tabular}
\caption{All examples of filtering a noisy command while removing important details of the target command.}
\label{tab:filtering-incorrect}
\end{table*}

\end{document}